\documentclass[preprint,2p,times]{elsarticle}

\usepackage{fullpage}
\usepackage{graphicx}
\usepackage{wrapfig}
\usepackage[english]{babel}
\usepackage[center]{caption}
\usepackage{mathtools}
\usepackage{mathptmx}    
\usepackage{latexsym}           
\usepackage{url}
\usepackage{multirow}
\usepackage{amssymb}
\usepackage{capt-of, blindtext}
\usepackage{pdfpages}
\usepackage{setspace}
\usepackage{lineno}
\usepackage{algorithmic}
\usepackage{algorithm}

\journal{}

\begin{document}

\begin{frontmatter}

	\title{The Dutch's Real World Financial Institute: Introducing Quantum-Like Bayesian Networks as an Alternative Model to deal with Uncertainty}

	\author[{label1}]{Catarina Moreira} \ead{cam74@le.ac.uk}
	\author[label2]{Emmanuel Haven}\ead{ehaven@mun.ca}
	\author[label1]{Sandro Sozzo}\ead{ss831@le.ac.uk}
	\author[label3]{Andreas Wichert} \ead{andreas.wichert@tecnico.ulisboa.pt}
	\address[label1]{School of Business and Research Centre IQSCS, University of Leicester, University Road, LE1 7RH Leicester, United Kingdom}
	\address[label2]{Faculty of Business Administration, Memorial University, 230 Elizabeth Ave, St. John's, NL A1C 5S7, Canada}
	\address[label3]{Instituto Superior T\'{e}cnico, INESC-ID, Av. Professor Cavaco Silva, 2744-016 Porto Salvo, Portugal}

	\begin{abstract}  
	
In this work, we analyse and model a real life financial loan application belonging to a 
sample
 bank in the Netherlands. The log is robust in terms of data, containing a total of 262 200 event logs, belonging to 13 087 different credit applications. The dataset is heterogeneous and consists of a mixture of computer generated automatic processes and manual human tasks. The goal is to 
 work out 
  a decision model, 
  which represents the underlying tasks that make up the loan application service,
   and to assess potential areas of improvement of the institution's internal processes. 
		To this end
		we study the impact of incomplete event logs for the extraction and analysis of business processes. It is quite common that event logs are incomplete with several amounts of missing information (for instance, workers forget to register their tasks). Absence of data is translated into a drastic
		decrease of precision and compromises the decision models, leading to biased and unrepresentative results. 

		We investigate how classical probabilistic models are affected by incomplete event logs and we explore quantum-like probabilistic inferences as an alternative mathematical model to classical probability. This work represents a first step towards 
systematic investigation of the impact of quantum interference in a real life large scale decision scenario. The results obtained in this study indicate that,	
 under high levels of uncertainty, the quantum-like models generate quantum interference terms, which allow an additional non-linear parameterisation of the data. Experimental results attest the efficiency of the quantum-like Bayesian networks, since the application of interference terms is able to reduce the error percentage of inferences performed over quantum-like models when compared to inferences produced by classical models.
	
		The 
		present study can open the way 
		towards the investigation and applicability of quantum-like models to financial economics, such as the analysis of a group of securities in portfolios or even risk management in banks.  
			
	\end{abstract} 


\end{frontmatter}

\doublespace

\section{Introduction}

In recent years, we have witnessed a vast increase in information. Given that the price of storage devices has been decreasing throughout the years, storing millions of records of information has become a common and affordable task in current enterprises / institutions. These large amounts of data pose serious difficulties in the extraction of valuable information, and the analysis of these datasets has become an extremely complex task. Enterprises often do not have control of the underlying processes that make up their products or services. This translates in workflow sequences with several redundant tasks, which play a crucial role in increasing the amount of expenses of a company and delays in the delivery of a final product / service to a client. 

In this work, we have the challenge to model a real life financial event log of a loan application belonging to a
sample
 bank in the Netherlands. The log is robust in terms of data, containing a total of 262 200 event logs, belonging to 13 087 credit applications. The only information known is that a customer selects a certain amount of money and submits her / his request to the bank's web platform. Some automatic tasks are triggered and 
 one can verify
  if an application is eligible for credit. The underlying tasks of this loan application are heterogeneous and consists of a mixture of computer generated processes and manual human tasks. The identification of the underlying processes that lead to a product / service is a very important task and an active research field in the scientific community, more specifically in the domain of {\it Business Process Management}~\citep{Aalst13}. 

\subsection{Business Process Management}

Defined as the set of techniques responsible for the optimisation of a company's business processes, {\it Business Process Management} promises the automatic detection of redundant tasks, cycles or unprofitable sequences of events, leading to an increase 
in 
the company's productivity, efficiency and a reduction of operational costs. Under these circumstances, a business process is understood as a collection of tasks that are linked and executed in a sequence until they result in a product or a service delivered to a client~\cite{Aalst04,Weske12}.   

One of the techniques used in Business Process Management (and which will be the focus of this work) is {\it Process Mining}. Process mining is a technique that enables the automatic analysis of business processes based on event logs. Instead of designing a workflow, process mining consists in gathering the information of the tasks that take place during the workflow process and storing that data in structured formats called {\it event logs}~\cite{Aalst04workflow}. While gathering this information, it is assumed that (1) each event refers to a task in the business process, (2) each event is associated 
with
 an instance of the workflow and (3) since the events are stored by their execution time, it is assumed that they are sorted~\citep{Aalst14}. This means that the ordering of the activities can be described by causal relationships, suggesting that decision models capable of representing cause/effect relationships are suitable models for the representation and analysis of the company's workflows and business process. Probabilistic graphical models, such as Bayesian Networks, are examples of decision models which are capable of representing influences or causal relationships between events~\cite{koller09prob}.

\subsection{The Problem of Missing Data}

Event logs are the main source of data for the discovery of the business processes that make up a company. However, it is quite common that event logs are incomplete with several amounts of missing information (for instance, workers forget to register their tasks, system crashes, etc). Usually statistical methods are applied to the existing data, in order to create knowledge and overcome the missing data. However, most of the statistical methods require 
a complete dataset (or at least a dataset sufficiently robust) in order to perform accurate predictions~\citep{Kang07}. Absence of data is translated into a drastically decrease of precision and compromises the statistical model, leading to biased and unrepresentative results. This affects all fields ranging from genetics~\citep{Diamond13}, psychology~\citep{McArdle01}, medical research~\citep{Seaman11}, etc.

Missing data involves (or leads to) high levels of uncertainty. In corporations, although many tasks are automated (that is, they are executed by computers), there is also a significant human component in these tasks. This human work force is subjected to human judgment errors, which needs to make decisions under scenarios with high levels of uncertainty (either missing data, untrusted information, or simply decisions under pressure). Human judgment errors can lead to redundant tasks in companies or lead to more unnecessary and more complex sequences of tasks, causing additional operational costs to companies or inaccurate decisions~\citep{Kahneman82book}. Under the scope of human judgment errors, there is a large collection of works over the literature reporting experimental findings demonstrating that humans constantly violate the laws of classical probability theory and logic in decision scenarios under uncertainty, leading to a set of decision paradoxes and fallacies~\cite{Tversky74,Tversky81,Tversky86,Kahneman79,Tversky92,Shafir92}.

\subsection{Quantum Cognition}

In order to accommodate decision paradoxes, a new discipline emerged in the last couple of decades, which is called {\it Quantum cognition}. Quantum cognition has emerged as a research field that aims to build cognitive models using the mathematical principles of quantum mechanics. Given that classical probability theory is very rigid in the sense that it poses many constraints and assumptions (single trajectory principle, obeys set theory, etc.), it becomes too limited (or even impossible) to provide simple models that can capture human judgments and decisions since people are constantly violating the laws of logic and probability theory~\citep{Busemeyer15,Busemeyer14,Aerts14}. 

Under a classical setting, probabilistic inferences are computed using the law of total probability. Let $A$ be a random variable defined by real numbers and contained in a sample space $\omega$, and let $B_i$ with $i = 1, \dots N$ be a partition of the same sample space, then the classical law of total probability is:
\[ Pr( A ) = \sum_{i = 1}^N Pr( B_i ) Pr( A | B_i ) \] 
Quantum cognition differs from classical theory in the following way. In quantum cognition, probabilities are defined by complex numbers, instead of real numbers, and are called amplitudes. In this work, we address to complex amplitudes by the symbol $\psi$. Note that a complex number is a number that can be expressed in the form $z = a + ib$, where $a$ and $b$ are real numbers and $i$ corresponds to the imaginary part, such that $i^2 = -1$. Alternatively, a complex number can be described in the form $z = \left| r \right| e^{i\theta}$, where $|r| = \sqrt{a^2 + b^2}$. The $e^{i\theta}$ term is defined as the phase of the amplitude. These amplitudes are related to classical probability by taking the squared magnitude of these amplitudes through Born's rule~\citep{Deutsch99}. This is achieved by multiplying the amplitude with its complex conjugate (Equation~\ref{eq:born}).
\begin{equation}
Pr( A ) = \left| \sum_{i = 1}^N \psi( B_i ) \psi( A | B_i ) \right|^2 
\label{eq:born}
\end{equation}
A consequence of using Born's rule to define probabilities is the emergence of quantum interference effects, which are the heart of quantum cognition. If we expand Equation~\ref{eq:born}, we will end up with a quantum probability formula, which contains two terms: one that corresponds to the classical probability and another term that corresponds to the quantum interference effects~\citep{Busemeyer12book}.  
\begin{equation}
Pr( A ) =\sum_{i = 1}^N Pr( B_i ) Pr( A | B_i ) + interference
\end{equation}

By manipulating the quantum interference term, we can disturb the classical probability values through constructive interferences (when the interference term is positive) or destructive interferences (when the interference term is negative). One can also look at quantum probabilistic inferences as an additional layer to classical inferences that allows a non-linear parameterisation of the data. Our hypothesis is that one can take advantage of this additional parametric layer and use it to improve the results of decision models in Business Process Management, which suffer from high levels of uncertainty due to the negative impacts of missing data. This would lead to more robust decision scenarios that could help reduce operational costs in companies by reducing insignificant tasks and consequently improving the service delivery times to clients.

So far, the literature shows that quantum cognitive models are able to accommodate many paradoxical situations in a general and straightforward framework~\cite{Aerts09,Aerts16,Busemeyer15trends}. There are also quantum predictive models that are able to predict the outcome of these decision scenarios with low percentage errors. However, current quantum cognitive models have been applied in very simple decision scenarios (for instance, the Prisoner's Dilemma), which can be modelled with at most two random variables~\cite{Yukalov10,Moreira16,Moreira17faces}. To the best of our knowledge, no quantum-like model has ever been applied in the context of a complex real life decision scenario, such as in Business Process Management.

\subsection{Main Contributions}

The applicability of quantum-like models in complex real life scenarios, such as medical decision-making problems or in economical / financial scenarios, is still an open research question in the literature and so far, to the best of our knowledge, no such studies have been conducted. For this reason, the purpose of this work is to give a first step towards this direction and test the effectiveness of quantum-like cognitive models in a real life financial scenario corresponding to a Dutch financial institute, which provides credit loans to its clients.

Since most of the times event logs are incomplete and lack large amounts of data, the main contribution of this work is the study of the impacts of missing data in the reconstruction of the institution's business processes. We investigate how classical probabilistic models are affected by missing data and we explore quantum-like probabilistic inferences as alternative mathematical models to classical probability models.


The main contributions of this work are the following:
\begin{enumerate}

\item {\bf Optimisation} of the institution's business processes by identifying and eliminating redundant tasks. This leads to an exponential drop in the {\it costs} and {\it times} that are involved in the loan application.

\item {\bf Extraction of Decision Model} representative of a reduced and optimised loan application of the institution's credit applications.

\item {\bf Dealing with Missing Data } by exploring the impact of two different probabilistic inference frameworks (one based on classical probability theory and another based quantum theory).

\end{enumerate} 

\subsection{Organisation}

Given the complexity of the problem, this work is segmented and organised in the following topics:

\begin{itemize}

\item Processing of the event log by discovering the underlaying information that makes up the event log and making sense of the relevance of this information for the construction of the business process (Section~\ref{sec:proc}).

\item Extraction of the institution's business process by extracting the sequence of tasks involved in each loan application from the financial institute event log and by detecting redundant and misconducted tasks (Section~\ref{sec:extract}); 

\item Construction of a decision model representative of the business process extracted. There are many options to be explored here. For this work, we opted for Bayesian Networks (Section~\ref{sec:BN});

\item Investigation of the impact of missing data in the event log for classical and non-classical probabilistic inferences. We explore alternative mathematical approaches to deal with uncertainty that are not based in classical probability theory. Again, there are many non-kolmogorovian probabilistic frameworks. For this work, and due to the recent successful application of quantum-like models~\citep{Busemeyer15}, we will investigate quantum-like probabilistic inferences (Section~\ref{sec:ql_inference}).

\end{itemize}

\section{Case Study: a Loan Application Bank in the Netherlands}

The event log that we use in this work is taken from a bank in the Netherlands and corresponds to a loan application, where customers request a certain amount of money. This dataset has been provided for the BPI Challenge in 2012 and is publicly available\footnote{BPI Challenge 2012 Dutch Financial Institute Dataset: \url{http://www.win.tue.nl/bpi/doku.php?id=2012:challenge}}. The only information that is given in that the application starts with a webpage that enables the submission of loan applications. A customer selects a certain amount of money and then submits his request. Then, the application performs some automatic tasks and checks if an application is eligible. If it is eligible, then the customer is sent an offer by mail. After this offer is received, it will be evaluated. In case of any missing information, the offer goes back to the client and is again evaluated until all the required information is gathered. A final evaluation is done to the application and it is approved~\citep{Bautista12}. 

It is also known that the process is composed of three different groups of processes. The first letter of each task corresponds to an identifier of the sub process it belongs to. The tasks that start with letter $A$ correspond to states of the application, which are computer automatic tasks. The tasks that start with letter $O$ correspond to offers, which are communicated to the client. It is not clear from the dataset if these tasks are automatically generated by the application or if they involve any human work. And the tasks that start with letter $W$ correspond to the work item belonging to the application and correspond to human tasks. 

\subsection{Processing the Event Log}\label{sec:proc}

The log consists of a structured file, which require a substantial amount of processing effort in order to identify and extract all relevant information for the analysis. In total, we identified 262 200 events, which are contained in 13 087 different loan applications. Each loan application is associated with some amount of money requested by the client. The summary of all different tasks extracted from the event log are discriminated throughout Tables~\ref{tab:Atasks} to~\ref{tab:Otasks}.

Table~\ref{tab:Atasks} summarises the computed automated tasks, $A\_$. These tasks correspond to the bank application and from it is understood from the data, the costumer triggers the initiation of the process by submitting some required amount of money. The root node of the entire application was identified as being $A\_SUBMITTED$. At this stage, we already identify some redundancy in the data since it seems that the processes $A\_SUBMITTED$ and $A\_PARTLYSUBMITTED$ always occur together and in sequence. This means that the bank application has an additional process that is unnecessarily consuming time and computer resources. However, we can only confirm this redundancy after analysing the graphical structure of the process (Section~\ref{sec:extract}). The same redundancy was found at the end of the application process. The three redundant end nodes identified were  $A\_APPROVED$, $A\_REGISTERED$ and $A\_ACTIVATED$. These three events always occur together interchangeably.

\begin{table}[h!]
\resizebox{\columnwidth}{!}{  
\begin{tabular}{l | c | l }
{\bf Event}					& Occurrences		& {\bf Description} \\
\hline
{\bf A\_SUBMITTED} 			& 13 087			& Initial states. All 13 087 cases recorded in the log file start with these events.\\
{\bf A\_PARTLYSUBMITTED} 		& 13 087			&These tasks correspond to the action of a client starting the submission 	\\ 
							&				& for a request of some amount of money to be loaned.	 				\\
\hline			
{\bf A\_PREACCEPTED} 			& 7 367			& The application has not been accepted, because it requires additional information. \\
							&				&	\\
\hline
 {\bf A\_ACCEPTED} 			& 5 113			& The application has been accepted and ready to go to the final stage. \\
							&				& However, it can still need some additional information from the client. \\
\hline
 {\bf A\_FNIALIZED} 			& 5 015			& The submitted application is fully accepted and ready for assessment.  \\
							&				&  \\
\hline
 {\bf A\_CANCELLED} 			& 2 807			& End states of an unsuccessful application process. \\
{ \bf A\_DECLINED} 			& 7 635			& Not clear what is the difference between them.\\
\hline
{\bf A\_APPROVED}				& 2 246			& Represent the end of a successful application process. \\
{\bf A\_REGISTERED} 			& 2 246 			& These three events always appear together interchangeably and \\
{\bf A\_ACTIVATED } 			& 2 246			& correspond to an approved loan application. \\
\hline 
\end{tabular}
}
\caption{System application tasks that were identified during the processing of the event log. Some redundant task were identified, but still not confirmed: \{ A\_SUBMITTED, A\_PARTLYSUBMITTED \} and \{ A\_APPROVED, A\_REGISTERED, A\_ACTIVATED \}~\citep{Bautista12}. }
\label{tab:Atasks}
\end{table}

Table~\ref{tab:Wtasks} summarises the tasks that correspond to manual Workers. The event log contains a time sequence information regarding these tasks, which can either be $START$, $SCHEDULE$ or $COMPLETE$. As the name indicates, $START$ corresponds to the beginning of a worker's task. When the worker has finished addressing the task, then the event state is changed to $COMPLETE$. Tasks that are postponed to some specified date (or time) are marked $SCHEDULE$.  For the analysis of the event log and for the extraction of the business process, we only considered the tasks that were in state $COMPLETE$.  Since these tasks are purely performed by humans, it is expected a lot of errors while conducting them. For instance, the task $W\_Change~contract~details$ exists on the system, however it has never been performed by any worker in the financial institute. 

\begin{table}[h!]
\resizebox{\columnwidth}{!}{  
	 \begin{tabular}{l | c | l}
        {\bf Event}					 		& Occurrences		& {\bf Description} \\
	\hline
	{\bf W\_Calling after sent offers} 			& 52 016			& Event triggered whenever there is an offer sent to a client	\\
										&				& 	\\
	\hline
 	{\bf W\_Assessing the application} 			& 20 809			& Evaluates whether the application is elicit for credit \\
										&				&     \\
	\hline
 	{\bf W\_Filling in information} 				& 54 850			&  Required after applications are pre accepted \\
										&				&  	\\
	\hline
 	{\bf W\_Fixing incoming lead} 				& 16 566			& Triggered by the initial application processes and  \\
										& 				& whenever a client did not fill all the required information			\\
	\hline
	{\bf W\_Calling to add missing information}	& 25 190			& Additional information needed after performing		\\
										&				& the application assessment			\\
	\hline								
	{\bf W\_Rate fraud}						& 664			&  Triggered after the assessment of the application,	\\
										&				&  it is investigated cases of suspicious fraud			\\
	\hline								
	{\bf W\_Change contract details}			& 0				&  Triggered when it is required a change in the contract	\\
										&				&		\\
	\hline
        \end{tabular}
} 
       \caption{Worker tasks that were identified during the processing of the event log. Workers tasks mean that these tasks are pure manual and performed by humans~\citep{Bautista12}.}
	\label{tab:Wtasks}
\end{table}

\begin{table}[b!]
\centering
\begin{tabular}{l | c | l }
	{\bf Event}				& Occurrences		& {\bf Description} \\
	\hline
	{\bf O\_CREATED} 			& 7 030			& Offer created for the client \\
	{\bf O\_SELECTED} 		& 7 030			& The client was selected to receive an offer	\\			
	{\bf O\_SENT} 				& 7 030			& Offer sent to the client	 \\
	\hline
 	{\bf O\_SENT BACK} 		& 3 454			& Client's response to the received offer	\\
							&				&     \\
	\hline
 	{\bf O\_ACCEPTED} 		& 2 243			& Corresponds to an end state of a successful offer	 \\
							&				& Both parties agree with the offer.	\\
	\hline
 	{\bf O\_CANCELLED} 		& 3 655			&  Corresponds to end states of an unsuccessful offer. \\
	{ \bf O\_DECLINED} 		& 802			& Either the client or the institution rejected the offers or 	\\
							&				& the offer was cancelled for some reasons		\\
	\hline
\end{tabular}
\caption{Tasks corresponding to Offers that were identified during the processing of the event log. These tasks are not fully known if they are conducted by works, by automatic application processes or by a mix of both~\citep{Bautista12}.}
\label{tab:Otasks}
\end{table}

Table~\ref{tab:Otasks} summarises the tasks that correspond to Offers. It is not clear from the dataset or from the information provided if these tasks correspond to human tasks or to applications tasks. We are guessing that they are a mix of both, but we will never know this with certainty. For what we understood from the process, whenever a loan application is elicit for credit, an offer is created and sent to the client. This offer can be sent back to the client, presumably if some changes are needed to the offer. It can also be accepted if the client accepts the offer or it can be cancelled or declined. Regarding these last two, it is not clear the difference between them, but it is supposed that an offer can be declined if the client or the institution rejects the offer.  Three possibly redundant tasks were also identified, given that they always appear together: $O\_CREATED$, $O\_SELECTED$ and $O\_SENT$. Again, this redundancy of tasks can contribute to a drop in the productivity of the service by consuming extra resources and time.

Summarising, the dataset contained a total of 262 200 events, which are contained in 13 087 different loan applications. We identified $24$ different events and several redundant events that could be subjected to some degree of optimisation. 

\subsection{Extracting a Business Model}\label{sec:extract}

\begin{figure}[h!]
\resizebox{\columnwidth}{!} {
\includegraphics{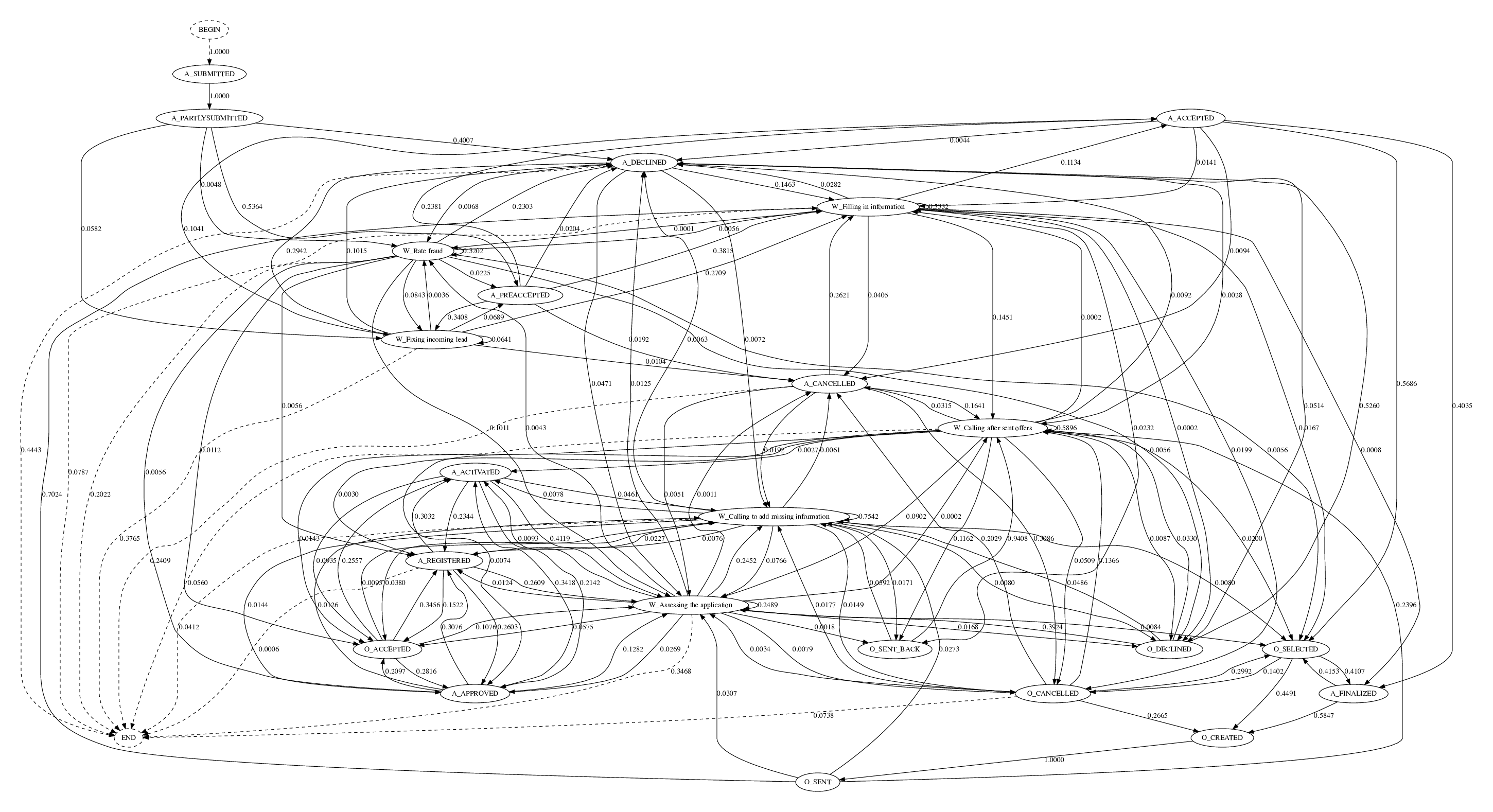}
}
\caption{Extracted business process from a Dutch's Financial Institute Dataset.}
\label{fig:markov_all}
\end{figure}

In a first step towards the understanding of the company's business processes, we generated a graphical model showing the sequence of all tasks that were conducted from the beginning of the loan application request, until its end (either with a successful application or with a rejection). The resulting plot shows a graphical structure where each node represents a task and each edge represents the probability of transiting from one task to another (Figure~\ref{fig:markov_all}). 

\begin{figure}[h!]
\resizebox{\columnwidth}{!} {
\includegraphics{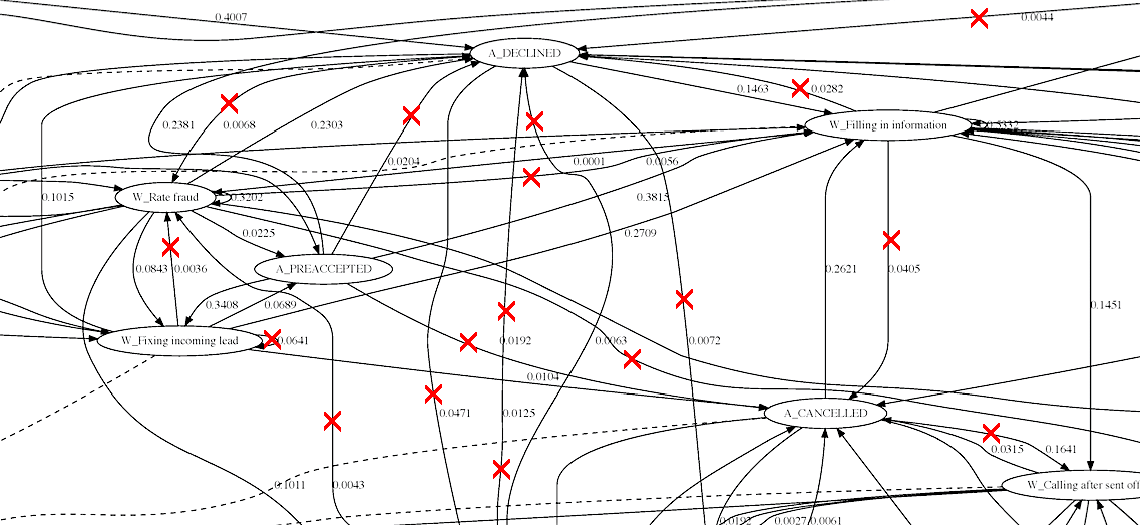}
}
\caption{Part of the business process extracted where we identify and remove very rare sequences of tasks. We consider that a sequence is rare if the probability of its occurrent is bellow $0.05$.}
\label{fig:markov_all_low_prob}
\end{figure}

To extract informational value out of the business process, we removed sequence of tasks that were very unlikely to occur. In other words, tasks that had very small transition probabilities. Consider Figure~\ref{fig:markov_all_low_prob}, which is a representation of a subset of the business process in Figure~\ref{fig:markov_all}. For instance, the probability of executing the sequence of tasks $A\_DECLINED \rightarrow W\_Rate Fraud$ is $0.0068$. Since the occurrence of the sequence of these tasks is very rare, one can ignore it and discard it from the analysis. In this work, we established that sequences of tasks with a transition probability bellow $0.05$ were not relevant to assess the value of the internal processes conducted in the company and, consequently, they were ignored. 

\subsection{Elimination of Redundant Tasks}

\begin{figure}[h!]
	\parbox{.33\columnwidth}{
	\centering
	\includegraphics[scale=0.37]{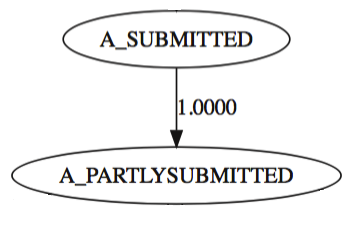}
	\caption{Redundancy found between events \{ A\_SUBMITTED, A\_PARTLYSUBMITTED \}.}
	\label{fig:init}
	}
	\hfill
	\parbox{.33\columnwidth}{
	\centering
	\includegraphics[scale=0.3]{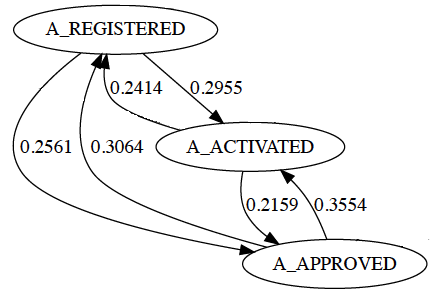}
	\caption{Redundancy found between events  \{ A\_APPROVED, A\_ACTIVATED, A\_REGISTERED \}.}
	\label{fig:end}	
	}
	\hfill
	\parbox{.33\columnwidth}{
	\centering
	\includegraphics[scale=0.35]{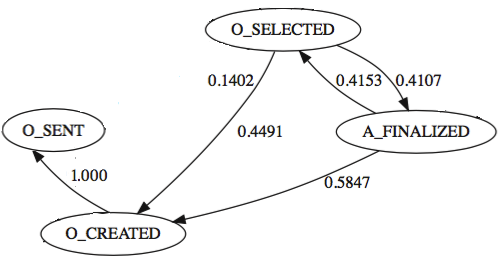}
	\caption{Redundancy found between events \{ O\_SELECTED, O\_CREATED, O\_SENT \}.}
	\label{fig:offers}	
	}
\end{figure}

When identifying the business processes from the event log, we suspected that there were several tasks, which were redundant and could be merged into a single task. Regarding the automatic processes, two sets of tasks were identified: \{ A\_SUBMITTED, A\_PARTLYSUBMITTED \} and \{ A\_APPROVED, A\_ACTIVATED, A\_REGISTERED \}. After extracting the causal relations and dependencies between events, we were able to confirm that in fact these tasks are redundant and contribute to an increase of operational costs and, consequently, on a decrease of productivity and efficiency. Considering Figure~\ref{fig:init}, we can see that after the root node $A\_SUBMITTED$, the node $A\_PARTLYSUBMITTED$ always occurs. To extract a more efficient business process out of the data, we will merge these two tasks into a single one and call it $A\_START\_APPLICATION$.

The same occurs for the ending processes (Figure~\ref{fig:end}). The dataset shows that before a credit is approved, these three nodes occur interchangeably. Again, they are consuming extra and unnecessary resources and in order to reduce the complexity of the model, we merged these tasks into a single one: \{ A\_APPROVED, A\_ACTIVATED, A\_REGISTERED \} $\rightarrow$ A\_CREDIT\_APPROVED.  

And, finally, in Figure~\ref{fig:offers}, the dataset shows that after an offer is created, the offer is always sent. Also, it seems that there are no rules in the application of the task $O\_SELECTED$. Almost half of the times it is triggered by the finalisation of the automatic process $A\_FINALIZED$. Some other times, it is the task $O\_SELECTED$ that triggers the $A\_FINALIZED$ task. This last transition makes no real sense, because first the automatic processes are conducted and only then, if they are successful, the manual tasks and offer tasks start. Given this order inconsistency, it seems that this task has been subjected to human intervention. It is straightforward that an offer cannot be done before the application process is finalised, so we know that $A\_FINALIZED$ precedes the creation of the offer. To avoid redundancy and inconsistencies, we decided to group the three tasks into a single one called $O\_CREATED AND SENT$, that is \{ O\_SELECTED, O\_CREATED, O\_SENT \} $\rightarrow$ O\_OFFER\_SENT.  Note that by removing these redundancies and unnecessary tasks, we were able to reduce the complexity of the business process from $24$ tasks to $18$.
 
\subsection{Elimination of Cycles}

The next step to optimise the business process is to eliminate cycles. This step plays an important role for two main reasons. First, it enables the discovery of cyclic sequences of tasks. Usually, these types of tasks are redundant and they contribute for the company's inefficiency. This translates again into a decrease in productivity and a vast increase in operational costs and production (or service delivery) time. Second, literature has reported the effectiveness of acyclic decision models as good approaches to model business processes and sequences of events~\citep{Bobek13}. A type of acyclic decision models that we are going to explore in this work are the Bayesian Networks~\cite{Pearl88}.

\begin{figure}[h!]
\centering
\includegraphics[scale=0.5]{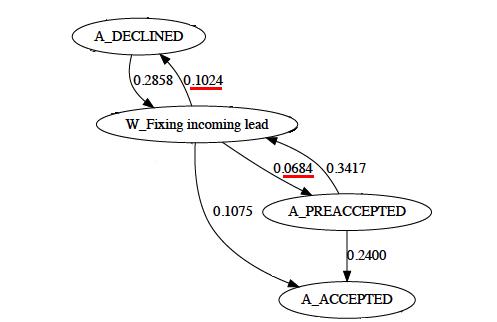}
\caption{Subprocess containing a transition with a cycle. The transition with the lowest probability was removed in order to guarantee an acyclic structure.}
\label{fig:markov_acyclic}
\end{figure}

These two reasons made us pursue the direction of eliminating cycles in the business process as a way to optimise the underlying processes that make up the financial institute. Figure~\ref{fig:markov_acyclic}, for instance, consists in a fragment of the business process, which contains cycles. One can easily notice that there could be human error between the transition of the manual task to the automatic task $W\_Fixing Incoming Lead \rightarrow A\_PREACCEPTED$ (which only contains a transition probability of $0.0684$) vs the opposite direction $A\_PREACCEPTED \rightarrow W\_Fixing~Incoming~Lead$ (which has a probability of $0.3417$). This actually makes some sense. Human worker's tasks are more subjected to human errors in contrast with pre-programmed computer automatised tasks. In these circumstances, we eliminate the cycle by simply deleting the edge with the lowest probability of occurrence. In Figure~\ref{fig:markov_acyclic}, the same reasoning can be made between tasks $W\_Fixing~Incoming~Lead$ and $A\_DECLINED$. 

\subsection{Final Network Structure}

Summarising, to extract a network structure representing the underlying processes that make up the financial institute, we proceeded in the following way:

\begin{enumerate}
\item Processing of the event log, identifying all tasks that were being conducted in the institute and determining the frequency of their occurrences. In the end, we identified 24 different tasks, contained in 262 200 events, which belonged to 13 087 different loan applications.

\item Extraction of a network structure, which initially was very complex to deal with due to the vast amount of transitions between tasks.

\item Optimisation of the network structure, which consisted in three main steps: (1) elimination of all edges with a transition probability bellow $0.05$, (2) identification and elimination of redundant tasks and (3) identification and elimination of cycles.

\end{enumerate}

\begin{figure}[h!]
\resizebox{\columnwidth}{!} {
\includegraphics{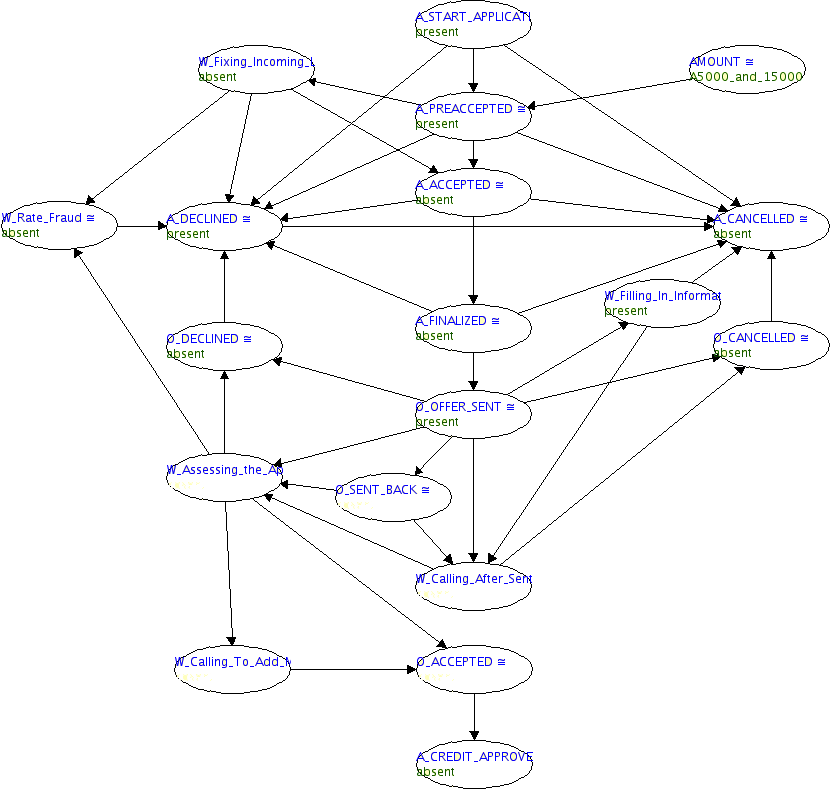}
}
\caption{Optimized and reduced acyclic network  structure extracted from the loan application bank event log.}
\label{fig:BN_all}
\end{figure}

In the end, we obtained a clear acyclic graphical structure (Figure~\ref{fig:BN_all}) representative of the business processes that makes up the financial institute from the beginning of a loan application until its end (either with a successful outcome or a denial). This structure is more clear and can now be analysed in therms of probabilistic inferences.

Given the acyclic structure of the network, the next step is to fill the corresponding conditional probability table, which show the probability distribution of a random variable  given its parents nodes. In the next section, we briefly explain how this was achieved.

\section{Learning the Conditional Probabilities}

The acyclic network structure that we obtained from the event log is called a Bayesian Network. Bayesian networks are probabilistic graphical models that are used to model decision scenarios and to make probabilistic inferences, that is, asking queries to the model and receiving answers in the form of probability values.

Under the realm of process mining, Bayesian Networks can represent activities as nodes (i.e. random variables) and the edges between activities can be seen as transitions between these tasks. From this structure, it is possible to automatically learn the conditional probability tables from a complete log of events using statistical models. Every node of the network is associated with a conditional probability tables, which specifies the probability distribution of a node, given its parents nodes.

\begin{figure}[h!]
\resizebox{\columnwidth}{!} {
\includegraphics{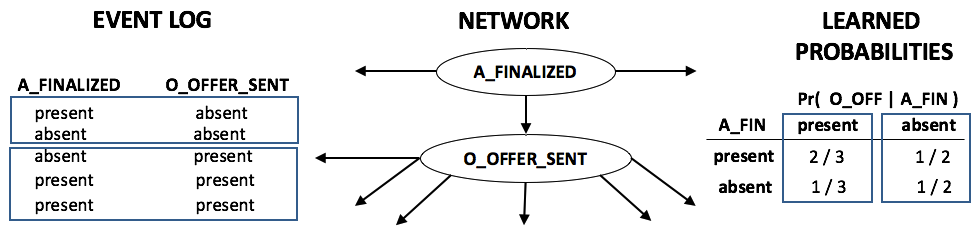}
}
\caption{Example of learning a conditional probability table from a complete dataset and a known network structure. The learning process consists in simply counting the number of occurrences of each assignment of the random variables and normalizing the final counts to obtain a probability value.}
\label{fig:learn_full}
\end{figure}

Having a complete network structure, estimation of the probabilities of a node given its parents nodes is straightforward. The financial institute provided a complete sample of their event log. When we have a known network structure and a full dataset, then the conditional probabilities of the network can be computed by simply counting how many times the conditioned variables occurred in the dataset. For instance, in the example in Figure~\ref{fig:learn_full}, the variable $O\_OFFER\_SENT$ has one single parent node, $A\_FINALIZED$. Both variables are binary and can represent the $presence$ or $absence$ of the event: if the task $A\_FINALIZED$ has been executed, then it is $present$, otherwise it is $absent$ from the application form. 

Using the example in Figure~\ref{fig:learn_full}, the learning process of a conditional probability table from a complete dataset with a known network structure simply consists in counting the number of occurrences of each assignment of the random variables and normalising the final counts to obtain a probability value. When the variable $O\_OFFER\_SENT$ has the value $present$ , there are $2$ out of $3$ entries in the dataset where its parent variable also occurs (probability of $0.67$) and $1$ out of $3$ entries where it does not (with probability $0.33$). In the same way, when $O\_OFFER\_SENT$ is $absent$, then we find that there is $1$ out of $2$ entries in the dataset where its parent variable is found to be $present$ and $absent$, leading to a probability of $0.5$.

One can see that the task of learning is very easy and straightforward in these circumstances, however, in most of the real world scenarios that is not the case. It is quite common that event logs are incomplete with several amounts of missing information (for instance, workers forget to register their tasks). Absence of data is translated into a drastically decrease of precision and compromises the statistical models, leading to biased and unrepresentative results. 

\begin{figure}[h!]
\resizebox{\columnwidth}{!} {
\includegraphics{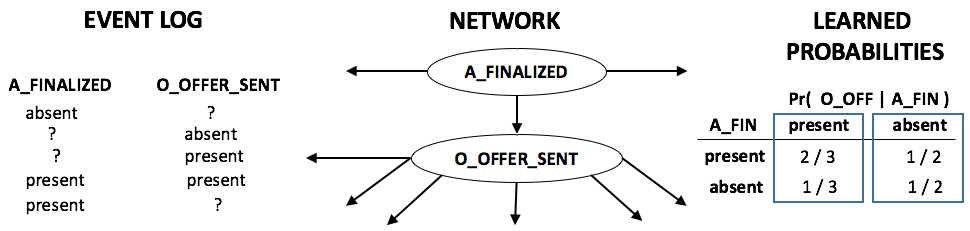}
}
\caption{Example of learning a conditional probability table from an incomplete dataset and a known network structure. The learning process consists in the application of statistical methods that assume that events are distributed according to a Gaussian distribution (the Expectation/Maximisation algorithm).}
\label{fig:learn_missing}
\end{figure}

For the study of this work, which consists in comparing the effectiveness of quantum-like probabilistic inferences with classical inferences, it is straightforward that for a complete dataset, it is understandable that the classical probabilistic inferences performed by the classical model will always be more representative of the data, because we are learning the data in a classical way. The interesting question to explore is, what is the impact of quantum-like probabilistic inferences when the dataset is not robust enough and suffers from a vast amount of missing information (which is actually quite common in real world scenarios). In this situation, the classical statistical models cannot generalise well and will lead to inaccurate results. To explore this condition, we randomly removed 70\% of the data from the event log and used a learning algorithm called {\it Expectation / Maximisation} to learn the conditional probability tables of the Bayesian Network~\citep{Dempster77}. Generally speaking, expectation / maximisation is a statistical method that assumes that data follows a Gaussian probability distribution. This way, the mean and the variance of the probability distribution can be estimated by only knowing a partial sample of the dataset. The details of this algorithm already fall out the scope of this paper, but the reader can refer to the book of~\citet{Bishop07} for further details. Figure~\ref{fig:learn_missing}, shows an example of what a dataset with missing data looks like and the final estimations of the conditional probability table learned with the expectation/maximisation algorithm.

\begin{figure}[h!]
	\parbox{.46\columnwidth}{
	\centering
	\includegraphics[scale=0.3]{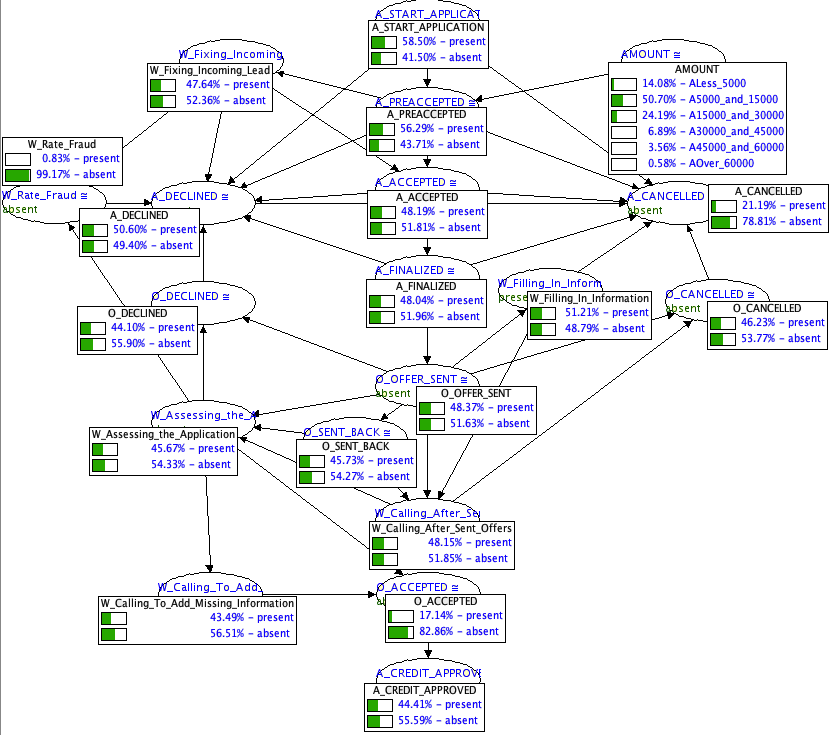}
	\caption{Resulting Bayesian network representing the business process of the financial institute with the conditional probability tables learned with $70\%$ of the data missing.}
	\label{fig:missing}
	}
	\hfill
	\parbox{.46\columnwidth}{
	\centering
	\includegraphics[scale=0.3]{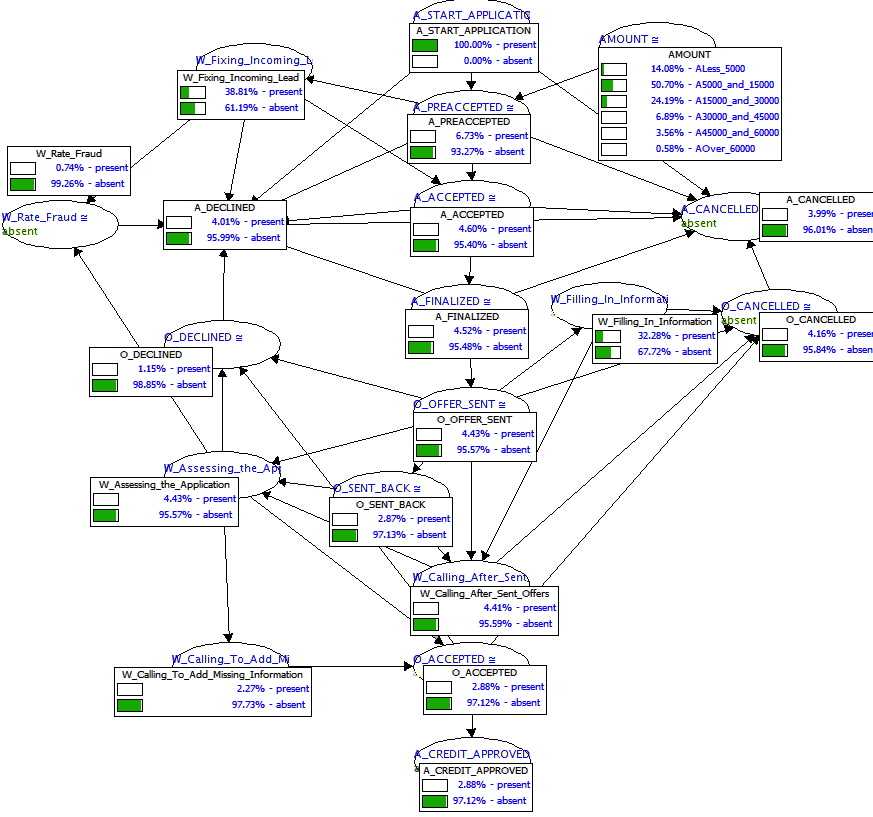}
	\caption{Resulting Bayesian network representing the business process of the financial institute with the conditional probability tables learned using the sull dataset.}
	\label{fig:complete}	
	}
\end{figure}

It is interesting to notice that the conditional probabilities learned using the incomplete dataset do not reveal much information about the underlying business processes of the institution. The conditional probability tables learned for most of the tasks has nearly a $50\%$ chance of either the task occurring or not. To give a more specific example, we can see that the probability of having a credit approved, $Pr( A\_CREDIT\_APPROVED )$, is $44.41\%$ in the Bayesian network learned with missing data (Figure~\ref{fig:missing}), in contrast with the $2.86\%$ obtained in the Bayesian network with the conditional probability tables learned using the full dataset (Figure~\ref{fig:complete}). 

Finishing the learning phase, we ended up with two classical Bayesian networks: one for the missing data and another one for the full data represented in Figures~\ref{fig:missing} and~\ref{fig:complete}, respectively. The Bayesian network in Figure~\ref{fig:complete} is our control network and will be used for evaluation purposes. It's conditional probability tables were learned using the full event log. On the other hand, the Bayesian Network on Figure~\ref{fig:missing} is the one that will be used to compare classical inferences over quantum-like inferences and its conditional probability tables were learned using the same event log, but with $70\%$ of its data randomly missing, this way introducing a high degree of uncertainty in the data.

At this stage one could be arguing about the effectiveness and applicability of Bayesian networks as appropriate decision models for process mining. Bayesian networks have already been used throughout the literature of business process management in many different scenarios~\citep{Bobek13}. Over the literature, Markov chains are the most commonly used models to represent business processes~\citep{Aalst13}. However, Bayesian networks provide a different decision-making analysis in the sense that they enable the specification of evidence variables. In other words, they provide the specification of some knowledge about the decision scenario. For example, suppose that the only thing that we know about the state of the application process is that a credit was approved. Then, we can ask the network what is the probability of a certain task occurring (for instance, $W\_Filling~In~Information$), given that we know that a credit was approved, $Pr( W\_Filling~In~Information | A\_CREDIT\_APPROVED)$. These types of inferences are unique to to Bayesian networks and provide an interesting type of analysis that is not commonly performed in such type of decision scenarios. What is even more interesting in this case study is that when we observe the state of the random variable \emph{A\_CREDIT\_APPROVED  = present}, then we know that the following events took place: \emph{A\_START\_APPLICATION} $\rightarrow$ \emph{A\_PREACCEPTED} $\rightarrow$ \emph{A\_ACCEPTED} $\rightarrow$ \emph{A\_FINALIZED} $\rightarrow$ \emph{O\_OFFER\_SENT} $\rightarrow$ \emph{W\_Filling\_In\_Information} $\rightarrow$ \emph{W\_Calling\_After\_Sent\_Offers} $\rightarrow$ \emph{W\_Assessing\_the\_application} $\rightarrow$ \emph{O\_ACCEPTED} $\rightarrow$ \emph{A\_CREDIT\_APPROVED}.

\begin{figure}[h!]
\centering
\includegraphics[scale=0.6]{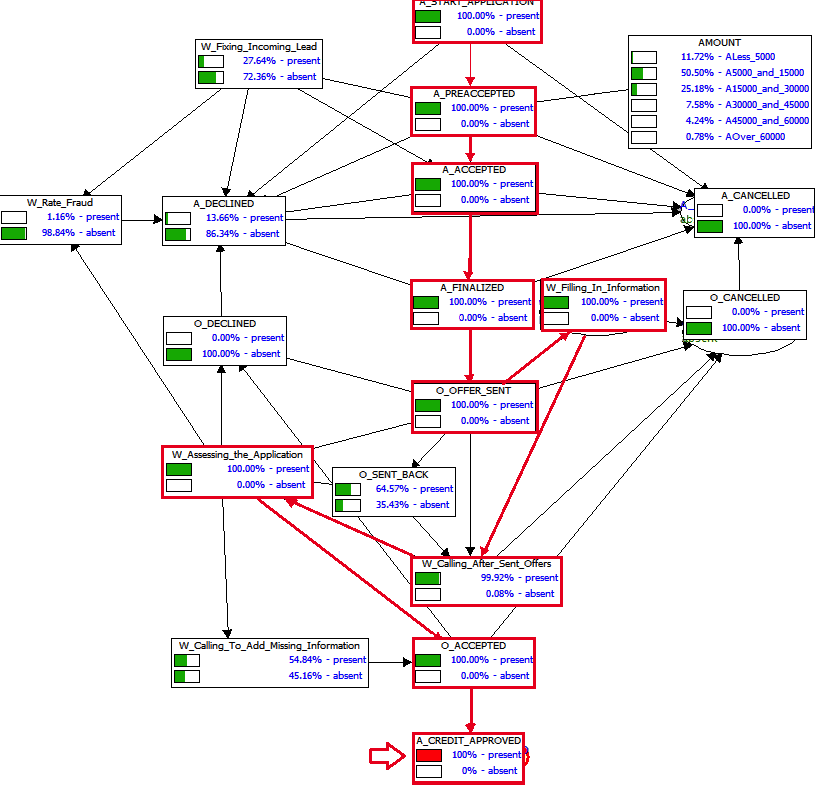}
	\caption{Impact of probabilistic inferences over Bayesian networks for process mining. Bayesian networks enables the specification of observed variables (evidence variables) and the specification of unobserved variables. In the figure, the only thing that was observed (piece of information provided) is that the variable $A\_CREDIT\_APPROVED$ was observed to be $present$. With this piece of information, we can know the entire workflow of the company with $100\%$ certainty.}
	\label{fig:credit_approved}	
\end{figure}
	
In the next section, we will formally present how to perform such types of probabilistic inferences both on classical and quantum-like Bayesian networks.

\section{Exact Inference in Classical and Quantum-Like Bayesian Network}\label{sec:BN}

Since the event logs of this financial institute are stored by their execution time, describing a causal sequence between events, we will explore the applicability and effectiveness of Quantum-Like Bayesian Networks~\cite{Moreira16} in the prediction of several events from the loan application process. A Quantum-Like Bayesian Network can be defined as an acyclic directed graph in which each node represents a random variable, each edge represents a direct influence from the source node to the target node and uses quantum probability amplitudes, which will be responsible for the emergence of quantum interference effects. Moreover, Bayesian Networks allows us to deal with uncertainty: each task can either be present or absent in the business process. Therefore, it is possible to perform special analysis that will enable the computation of the probability of some task of the business process occurring, given that we do not know which tasks have already been performed~\cite{Pearl88}.

\subsection{Classical Bayesian Networks}\label{sec:cl_inference}

A classical Bayesian Network can be defined by a directed acyclic graph structure in which each node represents a different random variable from a specific domain and each edge represents a direct influence from the source node to the target node. The graph represents independence relationships between variables and each node is associated with a conditional probability table which specifies a distribution over the values of a node given each possible joint assignment of values of its parents.  This idea of a node depending directly from its parent nodes is the core of Bayesian Networks. Once the values of the parents are known, no information relating directly or indirectly to its parents or other ancestors can influence the beliefs about it~\citep{koller09prob}.

\subsubsection{Classical Full Joint Distributions} 

In classical probability theory, the full joint distribution over a set of $N$ random variables $Pr(X_1, X_2, ..., X_N)$ corresponds to the probability distribution assigned to all of these random variables occurring together in the same sample space~\citep{koller09prob}.The full joint distribution of a Bayesian Network, where $X_i$ is the list of random variables and $Parents(X_i)$ corresponds to all parent nodes of $X_i$, is given by Equation~\ref{eq:joint}~\citep{russel10}: 

\begin{equation}
Pr( X_1, \dots, X_n ) = \prod_{i=1}^n  Pr( X_i | Parents(X_i) ) 
\label{eq:joint}
\end{equation}

\subsubsection{Classical Marginalization}

Given a query random variable $X$ and let $Y$ be the unobserved variables in the network, the marginal distribution of $X$ is simply the probability distribution of $X$ averaging over the information about $Y$. The marginal probability for discrete random variables, can be defined by Equation~\ref{eq:marginal_prob}. The summation is over all possible $y$, i.e., all possible combinations of values of the unobserved values $y$ of variable $Y$. The term $\alpha$ corresponds to a normalisation factor for the distribution $Pr( X )$~\citep{russel10}.

\begin{equation}
\begin{split}
Pr(X=x) = \alpha \sum_y Pr(X = x | Y = y) Pr(Y = y)\text{,~~~~~~~~~ where } \alpha = \frac{1}{ \sum_{x \in X} Pr(X = x) } 
\end{split}
\label{eq:marginal_prob}
\end{equation}

\subsection{Quantum-Like Bayesian Networks}\label{sec:ql_inference}

A quantum-like Bayesian Network can be defined by a directed acyclic graph structure in which each node represents a different quantum random variable and each edge represents a direct influence from the source node to the target node. The graph can represent independence relationships between variables, and each node is associated with a conditional probability table that specifies a distribution of quantum complex probability amplitudes over the values of a node given each possible joint assignment of values of its parents. In other words, a quantum-like Bayesian Network is defined in the same way as classical network with the difference that real probability values are replaced by complex probability amplitudes~\cite{Moreira16}

\subsubsection{Quantum-Like Full Joint Distribution}

The quantum-like full joint probability distribution can be defined in the same way as in a classical setting with two main differences: (1) the real probability values are replaced by complex probability amplitudes and (2) the probability value is given by applying the squared magnitude of a projection. In this sense, the quantum-like full joint complex probability amplitude distribution over a set of $N$ random variables $\psi(X_1, X_2, ..., X_N)$ corresponds to the probability distribution assigned to all of these random variables occurring together in a Hilbert space. Then, the full joint complex probability amplitude distribution of a quantum-like Bayesian Network is given by: 

\begin{equation}
\psi( X_1, \dots, X_N ) =  \prod_{j = 1}^N \psi ( X_j | Parents( X_j ) ) 
\label{eq:joint_q}
\end{equation}
Note that, in Equation~\ref{eq:joint_q}, $ X_i $ is the list of random variables (or nodes of the network), $Parents(X_i)$ corresponds to all parent nodes of $X_i$ and $\psi\left( X_i \right)$ is the complex probability amplitude associated with the random variable $X_i$. The probability value is extract by applying Born's rule, that is, by making the squared magnitude of the joint probability amplitude, $\psi \left( X_1, \dots, X_N \right)$:
\begin{equation}
Pr( X_1, \dots, X_N ) = \left| \psi( X_1, \dots, X_N ) \right|^2
\end{equation}

\subsubsection{Quantum-Like Marginalization}

The quantum-like marginalisation formula is the same as the classical one with two main differences: (1) the real probability values are replaced by complex probability amplitudes, (2) the probability is obtained by applying Born's rule to the equation. More formally, given a query random variable $X$ and let $Y$ be the unobserved variables in the network, the marginal distribution of $X$ is simply the amplitude probability distribution of $X$ averaging over the information about $Y$. The quantum-like marginal probability for discrete random variables, can be defined by Equation~\ref{eq:inference_q}. The summation is over all possible $y$, i.e., all possible combinations of values of the unobserved values $y$ of variable $Y$. The term $\gamma$ corresponds to a normalisation factor. Since the conditional probability tables used in Bayesian Networks are not unitary operators with the constraint of double stochasticity (like it is required in other works of the literature~\citep{busemeyer06,Busemeyer09}), we need to normalise the final scores. This normalisation is consistent with the notion of normalisation of wave functions used in Feynman's Path Diagrams.
In classical Bayesian inference, on the other hand, normalisation is performed due to the independence assumptions made in Bayes rule. 

\begin{equation}
\begin{split}
Pr(  X | e ) = \gamma \left|~ \sum_y \prod_{k=1}^N \psi( X_k | Parents(X_k), e, y ) ~\right| ^2
\end{split}
\label{eq:inference_q}
\end{equation}
Expanding Equation~\ref{eq:inference_q}, it will lead to the quantum marginalisation formula~\citep{Moreira14}, which is composed by two parts: one representing the classical probability and the other representing the quantum interference term (which corresponds to the emergence of destructive / constructive interference effects):
\begin{equation}
Pr(X | e) = \gamma \sum_{i = 1}^{|Y|}  \left| \prod_k^N  \psi( X_k | Parents(X_k), e, y = i )  \right| ^2 + 2 \cdot {\mathit Interference}
\label{eq:bn_inference_q}
\end{equation}
\begin{multline*}
{\mathit Interference} =\\  \sum_{i=1}^{|Y|-1} \sum_{j=i+1}^{|Y|}  \left| \prod_k^N \psi( X_k | Parents(X_k), e, y=i ) \right| \cdot \left| \prod_k^N  \psi( X_k | Parents(X_k), e, y= j ) \right| \cdot \cos( \theta_i - \theta_j )
\end{multline*}

Note that, in Equation~\ref{eq:bn_inference_q}, if one sets $(\theta_i - \theta_j)$ to $\pi/2$, then $\cos( \theta_i - \theta_j) = 0$. This means that the quantum interference term is canceled and the quantum-like Bayesian Network collapses to its classical counterpart. In other words, one can see the quantum-like Bayesian Network as a more general and abstract model of the classical network, since it represents both classical and quantum behaviour. Setting the angles to right angles means that all cosine similarities are either 0 or 1,  transforming a continuous-valued system to a Boolean-valued system. Moreover, if the Bayesian Network has $N$ binary random variables, we will end up with $2^N$ free quantum $\theta$ parameters, which is the size of the full joint probability distribution.

\begin{figure}[h!]
	\parbox{.45\columnwidth}{
	\centering
	\includegraphics[scale=0.4]{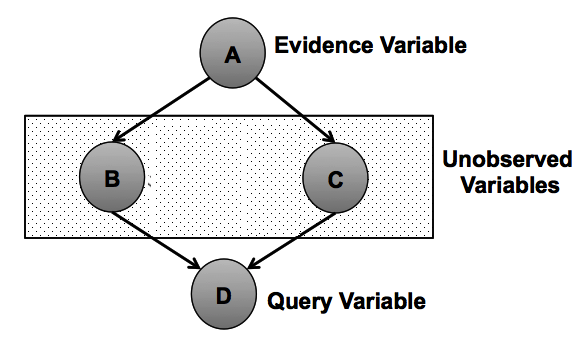}
	\caption{General example of a classical Bayesian network. Each node represent a random variable and each edge represents a direct influence from a source node to a target node. Each node is followed by a conditional probability table, which specifies the probaility distribution of a node given its parents.}
	\label{fig:classical_bn}
	}
	\hfill
	\parbox{.45\columnwidth}{
	\centering
	\includegraphics[scale=0.4]{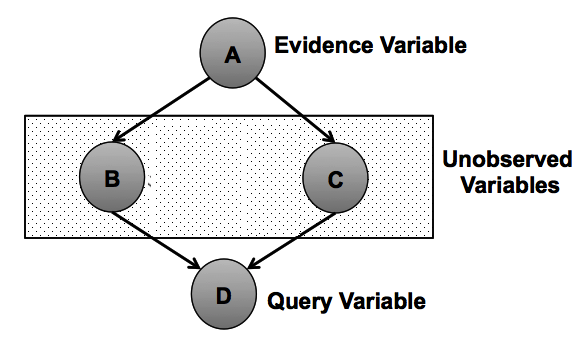}
	\caption{General example of a quantum-like Bayesian network. Each node represent a random variable and each edge represents a direct influence from a source node to a target node. Unobserved nodes can produce quantum interference effects, which can disturb the final probability outcomes.}
	\label{fig:quantum_bn}	
	}
\end{figure}

Formal methods to assign values to quantum interference terms is still an open research question, however some work has already been done towards that direction~\citep{Yukalov11,Moreira16,Moreira17faces}. In this work, we will use the heuristic developed in the work of~\citet{Moreira16} in order to set the quantum interference parameters.

\subsection{Quantum Interference Terms}

So far, we presented a general quantum-like Bayesian Network model, which performs quantum-like
probabilistic inferences. In the recent work of~\citet{Moreira16}, the authors proposed a similarity heuristic, which proved to be effective in paradoxical scenarios that were violating the Sure Thing Principle~\citep{savage54}. Note that an heuristic is simply a shortcut that generally provides good results in many situations (in this case, for violations to the Sure Thing Principle), but at the cost of occasionally not giving us very accurate results~\citep{Shah08}. 

\begin{figure}
\resizebox{\columnwidth}{!} {
\includegraphics{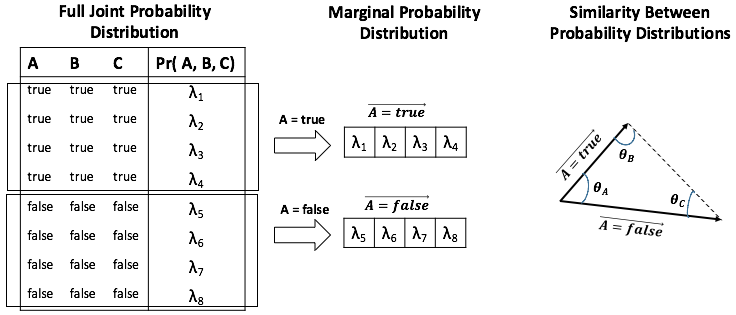}
}
\caption{Example of how to compute the similarity heuristic proposed in the previous work of~\citet{Moreira16}.  }
\label{fig:heuristic}
\end{figure}

Probabilistic inferences are computed by selecting from the full joint probability distribution the appropriate assignments. Following the example in Figure~\ref{fig:heuristic}, if we want to compute the probability of the random variable $A$ being $true$, $Pr( A = true )$, then one selects from the full joint probability distribution all the entries where $A = true$ and all entries where $A = false$. These entries correspond to the marginal probability distribution and, if we sum the values of the vectors and normalise them, we will end up with a classical probability answer to the query: 
\[ Pr( A = true ) = \alpha \sum_{i = 1}^N \lambda_i, \]
where $\alpha$ is the normalisation factor. If we add a quantum interference term to this formula, then we will end up with a quantum-like probability answer to the same query, $\gamma$ being the normalisation factor:
\[  Pr( A = true ) = \gamma \left( \sum_{i = 1}^N \lambda_i + 2 \sum_{i = 1}^{N-1} \sum_{j = i +1 }^N \sqrt{\lambda_i} \sqrt{\lambda_j} Cos \left( \theta_i - \theta_j \right) \right) \]

The quantum interference parameters $\theta$ are obtained by extracting the similarity values between the marginal distribution vectors. This is achieved by computing the cosine similarity between them, which is a widely used similarity function in information retrieval~\citep{Yates10}. Following Figure~\ref{fig:heuristic}, the cosine similarity will gives us three degrees of similarity between the vectors: $\theta_A$, $\theta_B$ and $\theta_C$. In the work of~\citet{Moreira16}, the authors created the similarity measure $\phi$, which is given by the ratio between the angles of the probability vectors
\[ \phi =  \frac{\left( \theta_C - \theta_B \right)}{\theta_A}   \]
Note that $\phi$ is obtained based on the marginal probability distribution of the data. It measures the relation between two probability values, because we are considering binary random variables, and nothing else.

Just like learning algorithms need to {\emph learn} the distribution of the data, in the quantum-like Bayesian network we also need to perform an analysis of the data in order to set the quantum interference terms. The way to set quantum interference terms is still an open research question in the literature. Usually, one needs to have prior knowledge of the outcome of a decision scenario and only then manually adjust the the quantum interference effects~\citep{Busemeyer09,busemeyer06,Khrennikov09}. This is feasible for very small and controlled decision scenarios, however when we move to large scale and complex decision scenarios with millions of quantum parameters to set, this approach is intractable. 

The similarity heuristic proposed by~\citet{Moreira16} requires the definition of some threshold values based on the similarity measure $\phi$. In their work, the authors were able to obtain proper thresholds to predict many different experiments, which were violating the Sure Thing Principle. In this work, since we are not dealing with violations to the Sure Thing Principle, it was required a preliminary analysis of the data in order to establish the thresholds (or boundaries) of the heuristic function. The function devised is represented in Equation~\ref{eq:heuristic}.

\begin{equation}
h_{\theta} = \left\{ 
  \begin{array}{l l}
	1.5408 			\text{~~~~~~~if } \phi < -2 \\
	1.5178  			\text{~~~~~~~if } \phi >= -2~~\& \&~~\phi <= 0\\
	\pi 				\text{~~~~~~~~~~~~~~~~if } \phi >= 0.15 \\
	0				\text{~~~~~~~~~~~~~~~~otherwise} \\
  \end{array} \right.
\label{eq:heuristic}
\end{equation}

It is important to note that both classical and quantum models have the same amount of information: they only use the marginal probability distribution. The difference relies in the fact that classical probability uses real numbers and quantum-like models use complex numbers, which will lead to the emergence of quantum interference effects that can be anything in a given range of values. That is also a reason why we need to specify these thresholds in the heuristic function, otherwise we would have no control over the interference terms. Appendix~\ref{sec:heuristic} presents in more detail how to compute the similarity heuristic for quantum-like inferences.

\section{Comparison between Classical and Quantum-Like Bayesian Networks}

After learning the conditional probabilities of the Bayesian network and after presenting the inference process in Bayesian networks (both classical and quantum-like), we will now proceed with a comparison of the probabilistic inferences obtained in both classical and quantum-like Bayesian networks in the scenario where $70\%$ of the data from the event log is missing.

We reinforce and remind that the fact that we randomly removed $70\%$ of the data is to simulate a real world situation. Although the full dataset was kindly provided by a finance institution, in real world scenarios, financial data suffers from the problem of incomplete data~\citep{Hasbrouck04}. That is why there is an increasing need in the usage of machine learning algorithms that try to learn a model that can generalise given some sample of data~\citep{McNelis05}. 

In order to compare classical probabilistic inferences with quantum-like inferences in the Bayesian network with missing data, we queried each variable of the Bayesian network and compared the outcome with a Bayesian network whose conditional probability tables were learned using the full data of the event log. 

The results of comparing the probabilistic inferences performed in a Bayesian network with classical and quantum-like inferences are detailed in Table~\ref{tab:results}.

\begin{table}[h!]
\resizebox{\columnwidth}{!} {
\begin{tabular}{ l | c c | c c | c | }
\cline{2-6}						& \multicolumn{4}{c | }{{\bf MISSING DATA BN}} &{\bf COMPLETE DATA BN}	\\
~												&  \multicolumn{2}{c }{{\bf Inferences}}		&\multicolumn{2}{c | }{ {\bf Error (\% )}}	& {\bf Inferences} 	 	\\
~												& {\bf Quantum}	& {\bf Classical}		& {\bf Quantum} & {\bf Classical}	& {\bf Classical (baseline)} \\
\hline
{\bf Pr( A\_PREACCEPTED = present ) }					& {\bf 0.1526}		& 0.3298				& {\bf 8.53}	& 26.25			& 0.0673	\\ 
{\bf Pr( A\_ACCEPTED = present ) }					& {\bf 0.1313}		& 0.3152				& {\bf 8.53}	& 26.92			& 0.0460	\\
{\bf Pr( A\_DECLINED	= present ) }					& {\bf 0.1293}		& 0.5325				& {\bf 8.92}	& 49.24			& 0.0401	\\
{\bf Pr( O\_SENT\_BACK = present ) }					& {\bf 0.0625}		& 0.4115				& {\bf 3.38}	& 38.28			& 0.0287	\\
{\bf Pr( O\_CANCELLED = present ) }					& {\bf 0.0740}		& 0.4160				& {\bf 3.24}	& 37.44			& 0.0416	\\
{\bf Pr( O\_DECLINED = present ) }						& {\bf 0.0584}		& 0.4070				& {\bf 4.69}	& 39.55			& 0.0115	\\
{\bf Pr( W\_Assessing\_the\_Application = present ) }		& {\bf 0.0740}		& 0.4160				& {\bf 3.24}	& 37.44			& 0.0443	\\
{\bf Pr( W\_Filling\_In\_Information = present ) } 			& {\bf 0.3981}		& 0.4706				& {\bf 7.53}	& 14.78			& 0.3228	\\
{\bf Pr( W\_Calling\_After\_Sent\_Offers = present ) } 		& {\bf 0.088}	 	& 0.4177				& {\bf 4.35}	& 37.36			& 0.0441	\\
{\bf Pr( W\_Calling\_To\_Add\_Missing\_Info = present ) }	& {\bf 0.3324}	 	& 0.4297				& {\bf 30.97}	& 40.70			& 0.0227	\\
\hline
\hline
{\bf Pr( A\_CREDIT\_APPROVED = present ) }			& {\bf 0.1674}		& {\bf 0.1674}			& {\bf 13.86}	& {\bf 13.86}		& 0.0288	\\
{\bf Pr( O\_ACCEPTED = present )	}					& {\bf 0.1674}		& {\bf 0.1674}			& {\bf 13.86}	& {\bf 13.86}		& 0.0288	\\
{\bf Pr( O\_OFFER\_SENT = present ) }					& {\bf 0.1014}		& {\bf 0.1014}			& {\bf 5.71}	& {\bf 5.71}		& 0.0443	\\		
{\bf Pr( W\_RATE\_FRAUD = present ) }					& {\bf 0.0082}		& {\bf 0.0082}			& {\bf 0.08}	& {\bf 0.08}		& 0.0074	\\
\hline
\hline
{\bf Pr( A\_FINALIZED = present ) }						& 0.1808			& {\bf 0.1786}			& 13.56		& {\bf 13.32}		& 0.0452	\\	
{\bf Pr( A\_CANCELLED = present ) }					& 0.1346			& {\bf 0.1260}			& 9.47		& {\bf 8.61}		& 0.0399	\\
{\bf Pr( W\_Fixing\_Incoming\_Lead = present )}			& 0.2900			& {\bf 0.4792}			& 9.82		& {\bf 9.11}		& 0.3881	\\	
\hline
\end{tabular}
}
\caption{Comparison between quantum-like and classical inferences over a Bayesian network learned using an incomplete dataset (with $70\%$ of missing data). The results show that quantum-like inferences achieved an average error of $8.35\%$ when compared to the $23.82\%$ error obtained in the classical inference. The column COMPLETE DATA BN represents the control network, which was learned using the full dataset.}
\label{tab:results}
\end{table}

The results show that the quantum-like inferences were able to adjust the probabilistic inferences of the classical network in scenarios with high levels of uncertainty (no variables observed). One can look at quantum-like probabilistic inferences as an additional layer to the classical inferences that allows a non-linear parameterisation of the data. 

It is interesting to notice that quantum-like inferences either outperform classical inferences or, in a worst case scenario, have the same performance as a classical network. This issue has already been noticed and pointed out in the previous studies of~\citet{Moreira16,Moreira14,Moreira15qi}. The queries performed over the random variables  $A\_FINALIZED$, $A\_CANCELLED$ and $W\_Fixing Incoming Lead$ were the ones with higher errors, but they had nearly the same performance as the classical network: the quantum-like model achieved an mean error of $10.95\%$ compared with the $10.35 \%$ mean error obtained in a classical setting. The general results show that the average error over the $19$ random variables, in scenarios where nothing is observed, for the quantum-like Bayesian network was $8.36\%$ compared to a $23.81\%$ error in the classical network. 

Although much more research needs to be done towards this direction, this study suggests that quantum-like inferences can be used as a way to complement inferences in classical models. This can have high impact in several domains where machine learning plays an important role (for instance, medical decision-making),

\subsection{Advantages and Disadvantages of Quantum-Like Bayesian Networks}

It is straightforward that quantum-like Bayesian networks suffer the same problem of the exponential increase of complexity (expressed as the dimension of the state space) as the classical Bayesian networks. Indeed, in what concerns the complexity of the inference problem, Bayesian networks (either classical or quantum-like) will always be \emph{NP-Hard}. This means that exact inference on Bayesian networks are part of a class of problems that are extremely hard for a computer to solve, because it takes an exponential number of computational steps to perform the computations. The hardness of the exact inference comes precisely in the computation of the full joint probability distribution, which takes at most $2^{N}-1$, computational steps assuming that all random variables of the network are binary, for $N$ being the number of nodes in the network. This gives a complexity of $O(2^N)$. If random variables are not binary, then the exact inference process becomes even worse with a complexity of $O(M^N)$, where $M$ is the number of assignments that the random variables can have. 

The initial analysis that we performed in this work enabled us to identify redundant tasks in the financial institute. The redundancy of these tasks lead to an increase in operational costs, and to a decrease in the productivity state of the company. With a preliminary analysis, we were able to decrease the number of tasks in the business process from $25$ events to $19$. In order to have some notion of the impact of this identification in the inference problem, if we used all tasks that were identified in the event log, we would end up with a full joint probability distribution of $6 \times 2^{23} = 50~331~648$ entries, which corresponds to the $AMOUNT$ random variable (which contains $6$ different assignments) and $23$ binary random variables (which contains $2^{23}$ different assignments). Under a classical setting, this is computationally intractable and in order to deal with these situation we could not use exact inference mechanisms. An alternative approach would be the usage of approximative inference methods, such as the belief propagation algorithm originally proposed by~\citet{Pearl88}. However, quantum-like versions of this algorithm have not been heavily explored in the literature. With the identification of the redundant tasks, we were able to reduce the state space to $6 \times 2^{19} = 3~145~728$ entries, which is already computationally tractable.

The quantum-like Bayesian network suffers from the same problem as the classical network in terms of the exponential increase of the full joint probability distribution, however, it also enables a new set of free parameters, which are the consequence of quantum interference effects. These interference effects can be seen as an additional non-linear parametrical layer that is added to classical inferences in order to refine probabilistic inferences. Of course, a preliminary analysis of the data needs to be performed in order to refine the boundaries that are required for the heuristic proposed in~\citet{Moreira16}. And the computation of these quantum interference effects can be performed in quadratic time with an addition of $m(m+1)/2 – m$ operations, where $m$ is the size of the marginal probability distribution. So, in the end, we lose a little bit of performance, but we are able to get a decision model that represents a decision scenario under high levels of uncertainty much better than a classical network. 

All simulations, the Bayesian networks and the code to perform classical and quantum-like inferences that we used in the experimental findings of this work were made freely available for researchers\footnote{\url{https://github.com/catarina-moreira/bpmn}}. 

\section{Conclusions}

In this work, we investigated how classical probabilistic models are affected by incomplete event logs and explored quantum-like probabilistic inferences as an alternative mathematical model to classical probability. We presented a pioneering study of the impact of quantum interference terms in a real life, large scale decision scenario. 

We analysed a dataset from a financial institute from the Netherlands concerning loan applications. We were able to discover the underlying processes that make up the institute's business process and we optimised the workflow by identifying redundant tasks and insignificant sequences of tasks. This procedure plays an important role for two main reasons. First, it increases the productivity and efficiency of the institute by reducing operational costs and by decreasing the total time needed to deliver a product / service to the client. Second, because it enables the modelling of optimised decision models that can assist financial chief officers in the process of decision making. We showed an example that, using the full dataset provided by the institute (and after performing the optimisation of the tasks), we were able to determine the workflow of a loan application with certainty only knowing that a credit was successfully approved to a client. However, access to full information is a luxury and also a rare situation. Data is usually missing or unreliable and, in the absence of data, statistical methods cannot come up with a general model representative of the data. For this reason, it is important the study of methods that are capable of dealing with incomplete datasets and uncertainty.

Quantum-like models are part of a recent research field called Quantum Cognition~\citep{Busemeyer12book}. They have been proved throughout the literature that they are capable of representing uncertainty in a more general way than classical models, due to the usage of quantum interference effects~\citep{busemeyer06,Khrennikov04quantum}. These interference effects can be seen as an additional non-linear parametrical layer that is added to classical inferences in order to refine probabilistic inferences. The drawback is that a preliminary analysis of the data needs to be performed in order to refine the boundaries that are required for the similarity parameter in the heuristic proposed in~\citet{Moreira16}. Also, the computation of these quantum interference effects can be performed in quadratic time. So, in the end, we lose a little bit of performance, but we gain in terms of accuracy. So far, quantum-like models have only been applied in very small and controlled experiments~\citep{Aerts16,Busemeyer09}. The study conducted in this work represents a first attempt to assess the effectiveness of quantum-like models in real life scenarios. From this work, we verified that under large and complex decision scenarios with high levels of uncertainty, quantum-like inferences were able to outperform classical inferences. 

This opens new insights towards the investigation of different and heterogeneous areas of research. Some of those areas pertain to financial economics, which include analysis of a group of securities in portfolios and risk management in banks using quantum-like inferences.

\section{Acknowledgements}

This work was supported by national funds through Funda\c{c}\~{a}o para a Ci\^{e}ncia e a Tecnologia (FCT) with reference UID/CEC/50021/2013. The funders had no role in study design, data collection and analysis, decision to publish, or preparation of the manuscript.

\bibliographystyle{model1b-num-names} 

\begin{thebibliography}{46}
\expandafter\ifx\csname natexlab\endcsname\relax\def\natexlab#1{#1}\fi
\providecommand{\bibinfo}[2]{#2}
\ifx\xfnm\relax \def\xfnm[#1]{\unskip,\space#1}\fi
\bibitem[{van~der Aalst(2004)}]{Aalst04}
\bibinfo{author}{W.~van~der Aalst}, \bibinfo{title}{Business process management
  desmystified: a tutorial on models, systems and sstandard workflow
  management}, in: \bibinfo{booktitle}{Lecture notes in Computer Science},
  \bibinfo{publisher}{Springer}, \bibinfo{year}{2004}.
\bibitem[{van~der Aalst(2013)}]{Aalst13}
\bibinfo{author}{W.~van~der Aalst}, \bibinfo{title}{Business process
  management: A comprehensive survey}, \bibinfo{journal}{ISRN Software
  Engineering Journal}  (\bibinfo{year}{2013}).
\bibitem[{van~der Aalst(2014)}]{Aalst14}
\bibinfo{author}{W.~van~der Aalst}, \bibinfo{title}{Process Mining: Discovery,
  Conformance and Enhancement of Business Processes},
  \bibinfo{publisher}{Springer}, \bibinfo{year}{2014}.
\bibitem[{van~der Aalst et~al.(2004)van~der Aalst, Weijters and
  Maruster}]{Aalst04workflow}
\bibinfo{author}{W.~van~der Aalst}, \bibinfo{author}{T.~Weijters},
  \bibinfo{author}{L.~Maruster}, \bibinfo{title}{Workflow mining: Discovering
  process models from event logs}, \bibinfo{journal}{IEEE Transactions on
  Knowledge and Data Engineering} \bibinfo{volume}{16} (\bibinfo{year}{2004})
  \bibinfo{pages}{1128--1142}.
\bibitem[{Aerts(2009)}]{Aerts09}
\bibinfo{author}{D.~Aerts}, \bibinfo{title}{Quantum structure in cognition},
  \bibinfo{journal}{Journal of Mathematical Psychology} \bibinfo{volume}{53}
  (\bibinfo{year}{2009}) \bibinfo{pages}{314--348}.
\bibitem[{Aerts(2014)}]{Aerts14}
\bibinfo{author}{D.~Aerts}, \bibinfo{title}{Quantum theory and human perception
  of the macro-world}, \bibinfo{journal}{Frontiers in Psychology}
  \bibinfo{volume}{5} (\bibinfo{year}{2014}) \bibinfo{pages}{1--19}.
\bibitem[{Aerts and Sozzo(2016)}]{Aerts16}
\bibinfo{author}{D.~Aerts}, \bibinfo{author}{S.~Sozzo}, \bibinfo{title}{From
  ambiguity aversion to a generalized expected utility. modeling preferences in
  a quantum probabilistic framework}, \bibinfo{journal}{Mathematical
  Psychology} \bibinfo{volume}{74} (\bibinfo{year}{2016})
  \bibinfo{pages}{117--127}.
\bibitem[{Arthur~Dempster and Rubin(1977)}]{Dempster77}
\bibinfo{author}{N.L. Arthur~Dempster}, \bibinfo{author}{D.B. Rubin},
  \bibinfo{title}{Maximum likelihood from incomplete data via the em
  algorithm}, \bibinfo{journal}{Journal of the Royal Statistical Society.
  Series B} \bibinfo{volume}{39} (\bibinfo{year}{1977}) \bibinfo{pages}{1--38}.
\bibitem[{Baeza-Yates and Ribeiro-Neto(2010)}]{Yates10}
\bibinfo{author}{R.~Baeza-Yates}, \bibinfo{author}{B.~Ribeiro-Neto},
  \bibinfo{title}{Modern Information Retrieval: The Concepts and Technology
  Behind Search}, \bibinfo{publisher}{Addison Wesley}, \bibinfo{year}{2010}.
\bibitem[{Bautista et~al.(2012)Bautista, Wangikar and Akbar}]{Bautista12}
\bibinfo{author}{A.~Bautista}, \bibinfo{author}{L.~Wangikar},
  \bibinfo{author}{S.~Akbar}, \bibinfo{title}{Process mining-driven
  optimization of a consumer loan approvals process: The bpic 2012 challenge
  case study}, in: \bibinfo{booktitle}{Lecture Notes in Business Information
  Processing}, \bibinfo{publisher}{Springer}, \bibinfo{year}{2012}.
\bibitem[{Bishop(2007)}]{Bishop07}
\bibinfo{author}{C.~Bishop}, \bibinfo{title}{Pattern Recognition and Machine
  Learning}, \bibinfo{publisher}{Springer}, \bibinfo{year}{2007}.
\bibitem[{Bobek et~al.(2013)Bobek, Baran, Kluza and Nalepa}]{Bobek13}
\bibinfo{author}{S.~Bobek}, \bibinfo{author}{M.~Baran},
  \bibinfo{author}{K.~Kluza}, \bibinfo{author}{G.~Nalepa},
  \bibinfo{title}{Application of bayesian networks to recommendations in
  business process modeling}, in: \bibinfo{booktitle}{Proceedings of the
  Central Europe Workshop}.
\bibitem[{Bruza et~al.(2015)Bruza, Wang and Busemeyer}]{Busemeyer15trends}
\bibinfo{author}{P.~Bruza}, \bibinfo{author}{Z.~Wang},
  \bibinfo{author}{J.~Busemeyer}, \bibinfo{title}{Quantum cognition: a new
  theoretical approach to psychology}, \bibinfo{journal}{Trends in Cognitive
  Sciences} \bibinfo{volume}{19} (\bibinfo{year}{2015})
  \bibinfo{pages}{383--393}.
\bibitem[{Busemeyer(2015)}]{Busemeyer15}
\bibinfo{author}{J.~Busemeyer}, \bibinfo{title}{Cognitive science contributions
  to decision science}, \bibinfo{journal}{Cognition} \bibinfo{volume}{135}
  (\bibinfo{year}{2015}) \bibinfo{pages}{43--46}.
\bibitem[{Busemeyer and Bruza(2012)}]{Busemeyer12book}
\bibinfo{author}{J.~Busemeyer}, \bibinfo{author}{P.~Bruza},
  \bibinfo{title}{Quantum Model of Cognition and Decision},
  \bibinfo{publisher}{Cambridge University Press}, \bibinfo{year}{2012}.
\bibitem[{Busemeyer and Wang(2014)}]{Busemeyer14}
\bibinfo{author}{J.~Busemeyer}, \bibinfo{author}{Z.~Wang},
  \bibinfo{title}{Quantum cognition: Key issues and discussion.},
  \bibinfo{journal}{Topics in Cognitive Science} \bibinfo{volume}{6}
  (\bibinfo{year}{2014}) \bibinfo{pages}{43--46}.
\bibitem[{Busemeyer et~al.(2006)Busemeyer, Wang and Townsend}]{busemeyer06}
\bibinfo{author}{J.~Busemeyer}, \bibinfo{author}{Z.~Wang},
  \bibinfo{author}{J.~Townsend}, \bibinfo{title}{Quantum dynamics of human
  decision making}, \bibinfo{journal}{Journal of Mathematical Psychology}
  \bibinfo{volume}{50} (\bibinfo{year}{2006}) \bibinfo{pages}{220--241}.
\bibitem[{Deutsch(1999)}]{Deutsch99}
\bibinfo{author}{D.~Deutsch}, \bibinfo{title}{Quantum theory of probability and
  decisions}, in: \bibinfo{booktitle}{Proceedings of the Royal Society A}.
\bibitem[{Diamond and Sekhon(2013)}]{Diamond13}
\bibinfo{author}{A.~Diamond}, \bibinfo{author}{J.~Sekhon},
  \bibinfo{title}{Genetic matching for estimating causal effects: A general
  multivariate matching method for achieving balance in observational studies},
  \bibinfo{journal}{Review of Economics and Statistics} \bibinfo{volume}{95}
  (\bibinfo{year}{2013}) \bibinfo{pages}{932--945}.
\bibitem[{Hasbrouck(2004)}]{Hasbrouck04}
\bibinfo{author}{J.~Hasbrouck}, \bibinfo{title}{Liquidity in the futures pits:
  Inferring market: Dynamics from incomplete data}, \bibinfo{journal}{Journal
  of Financial and Quantitative Analysis} \bibinfo{volume}{39}
  (\bibinfo{year}{2004}) \bibinfo{pages}{305--326}.
\bibitem[{Kahneman et~al.(1982)Kahneman, Slovic and Tversky}]{Kahneman82book}
\bibinfo{author}{D.~Kahneman}, \bibinfo{author}{P.~Slovic},
  \bibinfo{author}{A.~Tversky}, \bibinfo{title}{Judgment Under Uncertainty:
  Heuristics and Biases}, \bibinfo{publisher}{Cambridge University Press},
  \bibinfo{year}{1982}.
\bibitem[{Kahneman and Tversky(1979)}]{Kahneman79}
\bibinfo{author}{D.~Kahneman}, \bibinfo{author}{A.~Tversky},
  \bibinfo{title}{Prospect theory - an analysis of decision under risk},
  \bibinfo{journal}{Econometrica} \bibinfo{volume}{47} (\bibinfo{year}{1979})
  \bibinfo{pages}{263 -- 292}.
\bibitem[{Kang and Schafer(2007)}]{Kang07}
\bibinfo{author}{J.~Kang}, \bibinfo{author}{J.~Schafer},
  \bibinfo{title}{Demystifying double robustness: A comparison of alternative
  strategies for estimating a population mean from incomplete data},
  \bibinfo{journal}{Statistical Science} \bibinfo{volume}{22}
  (\bibinfo{year}{2007}) \bibinfo{pages}{523--539}.
\bibitem[{Khrennikov(2004)}]{Khrennikov04quantum}
\bibinfo{author}{A.~Khrennikov}, \bibinfo{title}{On quantum-like probabilistic
  structure of mental information}, \bibinfo{journal}{Journal of Open Systems
  and Information Dynamics} \bibinfo{volume}{11} (\bibinfo{year}{2004})
  \bibinfo{pages}{267--275}.
\bibitem[{Khrennikov(2009)}]{Khrennikov09}
\bibinfo{author}{A.~Khrennikov}, \bibinfo{title}{Description of composite
  quantum systems by means of classical random fields},
  \bibinfo{journal}{Foundations of Physics} \bibinfo{volume}{40}
  (\bibinfo{year}{2009}) \bibinfo{pages}{1051--1064}.
\bibitem[{Koller and Friedman(2009)}]{koller09prob}
\bibinfo{author}{D.~Koller}, \bibinfo{author}{N.~Friedman},
  \bibinfo{title}{Probabilistic Graphical Models: Principles and Techniques},
  \bibinfo{publisher}{The MIT Press}, \bibinfo{year}{2009}.
\bibitem[{McArdle and Hamagami(2001)}]{McArdle01}
\bibinfo{author}{J.~McArdle}, \bibinfo{author}{F.~Hamagami},
  \bibinfo{title}{Latent difference score structural models for linear dynamic
  analyses with incomplete longitudinal data}, \bibinfo{journal}{American
  Psychological Association} \bibinfo{volume}{24} (\bibinfo{year}{2001})
  \bibinfo{pages}{442}.
\bibitem[{McNelis(2005)}]{McNelis05}
\bibinfo{author}{P.~McNelis}, \bibinfo{title}{Neural Networks in Finance:
  Gaining Predictive Edge in the Market}, \bibinfo{publisher}{Elsevier Academic
  Press}, \bibinfo{year}{2005}.
\bibitem[{Moreira and Wichert(2014)}]{Moreira14}
\bibinfo{author}{C.~Moreira}, \bibinfo{author}{A.~Wichert},
  \bibinfo{title}{Interference effects in quantum belief networks},
  \bibinfo{journal}{Applied Soft Computing} \bibinfo{volume}{25}
  (\bibinfo{year}{2014}) \bibinfo{pages}{64--85}.
\bibitem[{Moreira and Wichert(2015)}]{Moreira15qi}
\bibinfo{author}{C.~Moreira}, \bibinfo{author}{A.~Wichert}, \bibinfo{title}{The
  relation between acausality and interference in quantum-like bayesian
  networks}, in: \bibinfo{booktitle}{Proceedings of the 9th International
  Conference on Quantum Interactions}.
\bibitem[{Moreira and Wichert(2016)}]{Moreira16}
\bibinfo{author}{C.~Moreira}, \bibinfo{author}{A.~Wichert},
  \bibinfo{title}{Quantum-like bayesian networks for modeling decision making},
  \bibinfo{journal}{Frontiers in Psychology} \bibinfo{volume}{7}
  (\bibinfo{year}{2016}).
\bibitem[{Moreira and Wichert(2017)}]{Moreira17faces}
\bibinfo{author}{C.~Moreira}, \bibinfo{author}{A.~Wichert},
  \bibinfo{title}{Exploring the relations between quantum-like bayesian
  networks and decision-making tasks with regard to face stimuli},
  \bibinfo{journal}{Journal of Mathematical Psychology} \bibinfo{volume}{78}
  (\bibinfo{year}{2017}) \bibinfo{pages}{86--95}.
\bibitem[{Pearl(1988)}]{Pearl88}
\bibinfo{author}{J.~Pearl}, \bibinfo{title}{Probabilistic Reasoning in
  Intelligent Systems: Networks of Plausible Inference},
  \bibinfo{publisher}{Morgan Kaufmann Publishers}, \bibinfo{year}{1988}.
\bibitem[{Pothos and Busemeyer(2009)}]{Busemeyer09}
\bibinfo{author}{E.~Pothos}, \bibinfo{author}{J.~Busemeyer}, \bibinfo{title}{A
  quantum probability explanation for violations of rational decision theory},
  \bibinfo{journal}{Proceedings of the Royal Society B} \bibinfo{volume}{276}
  (\bibinfo{year}{2009}) \bibinfo{pages}{2171--2178}.
\bibitem[{Russel and Norvig(2010)}]{russel10}
\bibinfo{author}{S.~Russel}, \bibinfo{author}{P.~Norvig},
  \bibinfo{title}{Artificial Intelligence: A Modern Approach},
  \bibinfo{publisher}{Pearson Education (3rd Edition)}, \bibinfo{year}{2010}.
\bibitem[{Savage(1954)}]{savage54}
\bibinfo{author}{L.~Savage}, \bibinfo{title}{The Foundations of Statistics},
  \bibinfo{publisher}{John Wiley}, \bibinfo{year}{1954}.
\bibitem[{Seaman and White(2011)}]{Seaman11}
\bibinfo{author}{S.~Seaman}, \bibinfo{author}{I.~White}, \bibinfo{title}{Review
  of inverse probability weighting for dealing with missing data},
  \bibinfo{journal}{Statistical Methods in Medical Research}
  \bibinfo{volume}{22} (\bibinfo{year}{2011}) \bibinfo{pages}{278--295}.
\bibitem[{Shafir and Tversky(1992)}]{Shafir92}
\bibinfo{author}{E.~Shafir}, \bibinfo{author}{A.~Tversky},
  \bibinfo{title}{Thinking through uncertainty: nonconsequential reasoning and
  choice}, \bibinfo{journal}{Cognitive Psychology} \bibinfo{volume}{24}
  (\bibinfo{year}{1992}) \bibinfo{pages}{449--474}.
\bibitem[{Shah and Oppenheimer(2008)}]{Shah08}
\bibinfo{author}{A.~Shah}, \bibinfo{author}{D.~Oppenheimer},
  \bibinfo{title}{Heuristics made easy: an effort-reduction framework.},
  \bibinfo{journal}{Psychological Bulletin} \bibinfo{volume}{134}
  (\bibinfo{year}{2008}) \bibinfo{pages}{207--222}.
\bibitem[{Tversky and Kahneman(1974)}]{Tversky74}
\bibinfo{author}{A.~Tversky}, \bibinfo{author}{D.~Kahneman},
  \bibinfo{title}{Judgment under uncertainty: Heuristics and biases},
  \bibinfo{journal}{Science} \bibinfo{volume}{185} (\bibinfo{year}{1974})
  \bibinfo{pages}{1124--1131}.
\bibitem[{Tversky and Kahneman(1981)}]{Tversky81}
\bibinfo{author}{A.~Tversky}, \bibinfo{author}{D.~Kahneman},
  \bibinfo{title}{The framing of decisions and the psychology of choice},
  \bibinfo{journal}{Science} \bibinfo{volume}{211} (\bibinfo{year}{1981})
  \bibinfo{pages}{453--458}.
\bibitem[{Tversky and Kahneman(1986)}]{Tversky86}
\bibinfo{author}{A.~Tversky}, \bibinfo{author}{D.~Kahneman},
  \bibinfo{title}{Rational choice and the framing of decisions},
  \bibinfo{journal}{The Journal of Business} \bibinfo{volume}{59}
  (\bibinfo{year}{1986}) \bibinfo{pages}{251--278}.
\bibitem[{Tversky and Shafir(1992)}]{Tversky92}
\bibinfo{author}{A.~Tversky}, \bibinfo{author}{E.~Shafir}, \bibinfo{title}{The
  disjunction effect in choice under uncertainty},
  \bibinfo{journal}{Psychological Science} \bibinfo{volume}{3}
  (\bibinfo{year}{1992}) \bibinfo{pages}{305--309}.
\bibitem[{Weske(2012)}]{Weske12}
\bibinfo{author}{M.~Weske}, \bibinfo{title}{Business Process Management:
  Concepts, Languages, Architectures}, \bibinfo{publisher}{Springer},
  \bibinfo{year}{2012}.
\bibitem[{Yukalov and Sornette(2010)}]{Yukalov10}
\bibinfo{author}{V.~Yukalov}, \bibinfo{author}{D.~Sornette},
  \bibinfo{title}{Entanglement production in quantum decision making},
  \bibinfo{journal}{Physics of Atomic Nuclei} \bibinfo{volume}{73}
  (\bibinfo{year}{2010}) \bibinfo{pages}{559--562}.
\bibitem[{Yukalov and Sornette(2011)}]{Yukalov11}
\bibinfo{author}{V.~Yukalov}, \bibinfo{author}{D.~Sornette},
  \bibinfo{title}{Decision theory with prospect interference and entanglement},
  \bibinfo{journal}{Theory and Decision} \bibinfo{volume}{70}
  (\bibinfo{year}{2011}) \bibinfo{pages}{283--328}.

\end{thebibliography}

\appendix

\section{Inferences in Quantum-Like Bayesian Networks}

The quantum-like Bayesian Network proposed in the previous work of~\citet{Moreira16} is built in a similar way as a classical network, with the difference that it uses quantum complex amplitudes to specify the conditional probability tables, instead of real probability values. As a consequence, the quantum-like Bayesian Network will give rise to quantum interference effects, which can act destructively or constructively if the interferences are negative or positive, respectively. 

Algorithm~\ref{alg:qlbn} describes the main steps to compute quantum-like inferences. Basically, a probabilistic inference consists in two major steps: the computation of the full joint probability distribution of the network and the computation of the marginal probability distribution with respect to the variable being queried.  

The algorithm starts by receiving a Bayesian Network represented by a set of {\it factors} specified by complex probability amplitudes instead of real probability values. A factor is a function that takes as input a set of random variables and returns all the assignments corresponding to that random variable. For instance, the full joint probability distribution of a network can be seen as a factor. The algorithm also receives a set of observed variables if some conditional probability is being queried. The random variable to be queried is also receive as input.

Given a Bayesian Network represented as set of factors, the algorithm first checks if there are any observed variables. More specifically, if the probabilistic inference is conditioned on some observed variable(s), then, for computational reasons, we set the values of the conditional probability tables, which are {\it not} consistent with the observed variables to $0$. By doing this, we are computing just the probabilities of the joint probability distribution that matter for the inference process, instead of computing the entire full joint probability distribution table.

\begin{algorithm} [h!]
\caption{Quantum-Like Bayesian Network}
\label{alg:qlbn}
\begin{algorithmic}[1]
\REQUIRE $\mathbb{F}$, factor structure  \\
		~~~~~~~~~~$ObservedVars$, list of observed variables,  \\
		~~~~~~~~~~$QueryVar$, identifier of the variable to be queried, \\
		~~~~~\\ 
\ENSURE Factor $Q$, corresponding to the quantum inferences, \\
	~~~~~~~~~Factor $C$, corresponding to the classical inferences \\
~~\\
\STATE /* A factor is a structure containing three lists: \\
~~~~$var$, corresponds to an identifier of a random variable. It also contains the list of the parent vars. \\
~~~~$card$, corresponds to the cardinality of each random variable in var. \\
~~~~$val$, corresponds to the respective conditional probability table. */ \\
~~\\
\STATE $Q \leftarrow struct('var', QueryVar , 'card', 2, val, \{ \});$~~~~~~~~~// initialise output factor structure for quantum network \\
\STATE $C \leftarrow struct(var, QueryVar, 'card', 2, val, \{ \});$~~~~~~~~~// initialise output factor structure for classical network \\
~~\\
\STATE // Observe evidence: set to 0 all factors in $\mathbb{F}$ that do not correspond to the evidence variables \\
\STATE $\mathbb{F} \leftarrow ObserveEvidence( \mathbb{F}, ObservedVars);$\\
~~\\
\STATE // Compute the Full Joint Probability Distribution of the Network: \\
\STATE \[ \psi( X_1, \dots, X_N ) =  \prod_{j = 1}^N \psi ( X_j | Parents( X_j ) )  \]
\STATE $Joint \leftarrow ComputeFullJointDistribution( \mathbb{F} );$ \\
~~\\
\STATE // Marginalise the full joint probability distribution. Select the positive and negative assignments of $QueryVar$: \\
\STATE  $\left[ PositiveProb, NegativeProb \right] \leftarrow FactorMarginalization(Joint, QueryVar);$ \\
~\\
\STATE // Compute classical probability factor by applying Equation~\ref{eq:marginal_prob}
\STATE $C.val \leftarrow ComputeClassicalProb( PositiveProb, NegativeProb  );$ \\
~\\
\STATE // Compute quantum probability factor according to Algorithm~\ref{alg:interf}\\
\STATE $Q.val \leftarrow ComputeQuantumProb( PositiveProb, NegativeProb  );$ \\
~\\
\RETURN $\left[ Q, C \right]$;

\end{algorithmic}
\end{algorithm}

Next, we compute the full joint probability distribution. This corresponds to the application of the full joint probability distribution formula described in Equation~\ref{eq:joint_q}. Basically, this function performs the product for each assignment of all random variables of the network. One needs to guarantee that the full joint probability distribution obeys to the normalisation axiom, making all entries of distribution sum to one. 

Having the full joint distribution factor, we can perform the probabilistic inference by computing the classical marginal probability distribution and the quantum interference term. The function {\it FactorMarginalization} corresponds to the selection of all entries of the full joint probability distribution that match the query variable and the evidence variables (if given). It returns two vectors: (1) one corresponding to the entries of the full joint probability where the query variable is observed to occur (we address these probabilities as $PositiveProb$), and (2) another one corresponding to the entries of the full joint probability where the query variable is observed to not occur ($NegativeProb$). The classical probability corresponds to a normalised summation of these vectors.

Having the vectors with the positive and negative probabilities resulting from the marginalisation process, we can also compute the quantum-like probabilities (Algorithm~\ref{alg:interf}). The quantum interference formula in Equation~\ref{eq:bn_inference_q} is given by set of two summations over the marginal probability vector. Due to normalisation purposes, we will need to compute the quantum interference term corresponding both to the positive and negative probability measures (when the query variable occurs and not occurs). The quantum interference parameter $\theta$ is computed according to the similarity heuristic and will be addressed with more detail in Section~\ref{sec:heuristic} of this Appendix. 

\begin{algorithm} [h!]
\caption{ComputeQuantumProbability}
\label{alg:interf}
\begin{algorithmic}[1]
\REQUIRE $PositiveProb$, vector of marginal probabilities when $QueryVar$ occurs,  \\
          ~~~~~~~~~~$NegativeProb$, vector of marginal probabilities when $QueryVar$ does not occur,  
~~ \\
		
\ENSURE List $Q$ with probabilistic inference using quantum theory \\
~~\\

\STATE $interference\_pos \leftarrow 0;$ \\
\STATE $length\_assign \leftarrow length(PositiveProb);$ \\
~~\\
\STATE // For all probability assignments, \\
\FOR{$i = 1; 1 i \leq length\_assign - 1; i = i + 1$}
\FOR{$j = i + 1; j \leq length\_assign; j = j + 1$}

\STATE // Compute the quantum interference parameter $\theta$ according to a given heuristic function \\
\STATE  $heurs \leftarrow SimilarityHeuristic( PosAssign, NegAssign )$\\
~~\\
\STATE // Apply quantum interference formula: \\
\STATE \[ \sum_{i=1}^{|Y|-1} \sum_{j=i+1}^{|Y|}  \left| \prod_k^N \psi( X_k | Parents(X_k), e, y=i ) \right| \cdot \left| \prod_k^N  \psi( X_k | Parents(X_k), e, y= j ) \right| \cdot \cos( \theta_i - \theta_j ); \] \\
\STATE // Compute the interference term related to the positive assignments \\
\STATE $interference\_pos \leftarrow interference\_pos + 2 PosAssign\left[ i \right] PosAssign\left[ j \right]~heurs $~~\\
\STATE ~~\\
\STATE // Compute the interference term related to the negative assignments (for normalisation) \\
\STATE $interference\_neg \leftarrow interference\_neg + 2 NegAssign\left[ i \right] NegAssign\left[ j \right]~heurs$~~\\
~~\\ 
\ENDFOR
\ENDFOR
~~\\ 
\STATE // Compute quantum-like probabilities: $classicalProb + interference$. \\  
\STATE $\alpha = (sum(PosAssign) + sum(NegAssign))^{-1}$ \\
\STATE $classicalProb \leftarrow \left[ \alpha~PosAssign, \alpha~NegAssign \right];$ \\
\STATE $probPos \leftarrow classicalProb[1] + interference\_pos;$ \\
\STATE $probNeg \leftarrow classicalProb[2] + interference\_neg;$ \\
~~\\
\STATE // Normalise the results in order to obtain a probability value \\
\STATE $\gamma \leftarrow (probPos + probNeg)^{-1}$ \\
\STATE $Q \leftarrow \left[ \gamma~probPos, \gamma~probNeg \right]$ \\
~~\\
\RETURN $Q$; \\

\end{algorithmic}
\end{algorithm}

\section{The Similarity Heuristic For Quantum Interference Effects}\label{sec:heuristic}

The goal of the similarity heuristic is to determine an angle between the probabilistic vectors associated with the marginalisation of the positive and negative assignments of the query variable. In other words, when performing a probabilistic inference from a full joint probability distribution table, we select from this table all probabilities that match the assignments of the query variable. If we sum these probabilities, we end up with a final classical probability inference. If we add an interference term to this classical inference, we will end up with a quantum-like inference. In this case, we can use these probability vectors to obtain additional information to compute the quantum interference parameters. The general idea of the similarity heuristic is to use the marginal probability distributions as probability vectors and measure their similarity through the law of cosines formula, which is a similarity measure well known in the Computer Science domain and it is widely used in Information Retrieval~\citep{Yates10}. According to this degree of similarity, we will apply a mapping function with an heuristically nature, which will output the value for the quantum interference parameter $\theta$ by taking into consideration a previous study of the probabilistic distribution of the data of several experiments reported over the literature. 

\begin{algorithm} [h!]
\caption{SimilarityHeuristic}
\label{alg:similarityHeurs}
\begin{algorithmic}[1]
\REQUIRE $PositiveProb$, vector of marginal probabilities when $QueryVar$ occurs,  \\
          ~~~~~~~~~~$NegativeProb$, vector of marginal probabilities when $QueryVar$ does not occur,  \\
	
\ENSURE $interf$, Quantum Interference term \\
~~\\
\STATE // Compute Euclidean distances between vectors
\STATE $norm_c \leftarrow norm( PosProb - NegProb, 2);$ \\
\STATE $norm_a \leftarrow norm( PosProb, 2);$ \\
\STATE $norm_b \leftarrow norm( PosNeg, 2);$ \\
~~\\
\STATE // Compute angles between vectors using the law of cosines
\STATE $\theta_a \leftarrow ACos( \frac{norm_b^2 - norm_a^2 + norm_c^2}{ 2~norm_c~norm_b});$ \\
\STATE $\theta_b \leftarrow ACos( \frac{norm_a^2 - norm_b^2 + norm_c^2}{2*norm_c*norm_a})$ \\
\STATE $\theta_c \leftarrow ACos( \frac{norm_a^2 + norm_b^2 - norm_c^2}{2*norm_a*norm_b});$ \\
~~\\
\STATE // Compute de similarity measure $\phi$ \\
\STATE $\phi \leftarrow \frac{ \theta_c }{ \theta_a } - \frac{ \theta_b  }{ \theta_a };$\\
~~\\ 
\STATE // Apply heuristic using the thresholds according to Equation~\ref{eq:heuristic} \\
\STATE $interf \leftarrow 0;$ \\
~~\\ 
 \IF{ $\phi < -2$  } 
\STATE $interf \leftarrow 1.5408;$\\
\ENDIF
~~\\ 
 \IF{ $\phi >= -2~~~\& \&~~~\phi <= 0$  } 
\STATE $interf \leftarrow 1.5178$\\
\ENDIF
~~\\ 
 \IF{ $\phi >= 0.15$  } 
\STATE $interf \leftarrow \pi$\\
\ENDIF
~~\\ 
\RETURN $Cos( interf )$; \\
\end{algorithmic}
\end{algorithm}

When performing quantum-like probabilistic inferences, two steps are required: (1) the computation of a quantum-like full joint probability distribution and (2) the computation of the quantum-like marginal distribution. The quantum superposition vector, comprising all possible events, is given by the quantum full joint probability distribution already presented in Equation~\ref{eq:joint_q}. 

Algorithm~\ref{alg:similarityHeurs} presents the pseudocode of the proposed heuristic. Given two vectors: (1) one corresponding to the entries of the full joint probability where the query variable is observed to occur (we address these probabilities as $PositiveProb$), and (2) another one corresponding to the entries of the full joint probability where the query variable is observed to not occur ($NegativeProb$). Then, one can compute the similarity heuristic in the following way.

First, one computes the euclidean distances between both vectors. Having the distances, one can use the law of cosines measure to determine the angles between all these vectors. With all this information, one can compute the similarity measure $\phi$ of the vectors and get the output of the quantum interference parameter. In the end, the algorithm returns the cosine of this value.

\end{document}